%% file: main.tex
\documentclass[letterpaper]{article} % DO NOT CHANGE THIS
\usepackage{aaai23} % DO NOT CHANGE THIS
\usepackage{times}  % DO NOT CHANGE THIS
\usepackage{helvet}  % DO NOT CHANGE THIS
\usepackage{courier}  % DO NOT CHANGE THIS
\usepackage[hyphens]{url}  % DO NOT CHANGE THIS
\usepackage{graphicx} % DO NOT CHANGE THIS
\usepackage{natbib}  % DO NOT CHANGE THIS AND DO NOT ADD ANY OPTIONS TO IT
\usepackage{caption} % DO NOT CHANGE THIS AND DO NOT ADD ANY OPTIONS TO IT
\usepackage{algorithm}
\usepackage{algorithmic}
\usepackage{amsfonts}
\usepackage{amsmath}
\usepackage{threeparttable}
\usepackage{booktabs}
\usepackage{subfig}

\usepackage{newfloat}
\usepackage{listings}
\DeclareCaptionStyle{ruled}{labelfont=normalfont,labelsep=colon,strut=off} % DO NOT CHANGE THIS
\floatstyle{ruled}
\newfloat{listing}{tb}{lst}{}
\floatname{listing}{Listing}
\title{GENNAPE: Towards Generalized Neural Architecture \\Performance Estimators}
\author{
    Keith G. Mills\textsuperscript{\rm 1,2}\thanks{Work done during an internship at Huawei.}, Fred X. Han\textsuperscript{\rm 2}, Jialin Zhang\textsuperscript{\rm 3}, Fabian Chudak\textsuperscript{\rm 2},\\ Ali Safari Mamaghani\textsuperscript{\rm 1}, Mohammad Salameh\textsuperscript{\rm 2}, Wei Lu\textsuperscript{\rm 2}, Shangling Jui\textsuperscript{\rm 3}, Di Niu\textsuperscript{\rm 1}
}
\affiliations{
    \textsuperscript{\rm 1}Department of Electrical and Computer Engineering, University of Alberta\\
    \textsuperscript{\rm 2}Huawei Technologies, Edmonton, Alberta, Canada\\
    \textsuperscript{\rm 3}Huawei Kirin Solution, Shanghai, China\\
    \{kgmills, safarima, dniu\}@ualberta.ca\\
    \{fred.xuefei.han1, fabian.chudak, mohammad.salameh, jui.shangling\}@huawei.com\\
    \{zhangjialin10, robin.luwei\}@hisilicon.com\\
}

\begin{document}

\maketitle

\begin{abstract}
\input{src/abstract}
\end{abstract}

\input{src/intro}
\input{src/related}
\input{src/family}
\input{src/method}
\input{src/results}
\input{src/conclusion}

\bibliography{references}

\clearpage 
\newpage
\input{src/supplementary}

\end{document}

%% file: src/abstract.tex
Predicting neural architecture performance is a challenging task and is crucial to neural architecture design and search. Existing approaches either rely on neural performance predictors which are limited to modeling architectures in a predefined design space involving specific sets of operators and connection rules, and cannot generalize to unseen architectures, or resort to Zero-Cost Proxies which are not always accurate. In this paper, we propose GENNAPE, a Generalized Neural Architecture Performance Estimator, which is pretrained on open neural architecture benchmarks, and aims to generalize to completely unseen architectures through combined innovations in network representation, contrastive pretraining, and a fuzzy clustering-based predictor ensemble. Specifically, GENNAPE represents a given neural network as a Computation Graph (CG) of atomic operations which can model an arbitrary architecture. It first learns a graph encoder via Contrastive Learning to encourage network separation by topological features, and then trains multiple predictor heads, which are soft-aggregated according to the fuzzy membership of a neural network. Experiments show that GENNAPE pretrained on NAS-Bench-101 can achieve superior transferability to 5 different public neural network benchmarks, including NAS-Bench-201, NAS-Bench-301, MobileNet and ResNet families under no or minimum fine-tuning. We further introduce 3 challenging newly labelled neural network benchmarks: HiAML, Inception and Two-Path, which can concentrate in narrow accuracy ranges. Extensive experiments show that GENNAPE can correctly discern high-performance architectures in these families. Finally, when paired with a search algorithm, GENNAPE can find architectures that improve accuracy while reducing FLOPs on three families.  

%% file: src/intro.tex
\section{Introduction}
\label{sec:intro}

Understanding and predicting the performance of neural networks is crucial to automated neural architecture design and neural architecture search (NAS). In fact, the amount of time and computational resources required by a NAS scheme is critically dependent on how evaluation is performed. Early methods~\cite{zoph2017NAS} rely on training individual architectures from scratch via intensive GPU usage. Weight-sharing methodologies like DARTS~\cite{liu2018DARTS} propose supernet models that superimpose the features of all architectures in a search space, substantially reducing the cost of search. DARTS further reduces this cost by searching for an architecture on a small image classification dataset like CIFAR-10~\cite{Krizhevsky09CIFAR} before transferring the found architecture to a larger benchmark dataset, like ImageNet~\cite{russakovsky2015imagenet}. In contrast, Once-for-All (OFA)~\cite{cai2020once} employs strategies to effectively pre-train a re-usable supernet %once 
on ImageNet prior to performing search. Newer methods have seen the introduction of Zero-Cost Proxies~\cite{abdelfattah2021zero} which estimate performance from forward pass gradients. 

Advancements in neural predictors~\cite{tang2020semi, wen2020neural} 
allow for faster inference time, which significantly lowers the architecture evaluation cost incurred in NAS. However, one major trade-off associated with using neural predictors is a lack of generalizability. The input to a neural predictor is usually a vector encoding of the architecture, whose scope is confined to the designated search space the predictor is meant to operate in. For example, to facilitate architecture search, Once-for-All, SemiNAS~\cite{luo2020semi} and BANANAS~\cite{white2019bananas} design specialized predictors, each performing in an individually defined search space. Such a restriction seriously limits the scope that a search algorithm~\cite{mills2021l2nas, mills2021exploring} can explore, regardless of how effective it is.

Furthermore, as a separate new predictor must be trained for each new search space created, there is an implied additional heavy cost associated with the laborious and resource-intensive task of labelling the performance of a suitable amount of neural architecture samples in this search space, since these architecture samples must be trained from scratch for evaluation. Ideally, a generalized neural predictor would not be subject to these constraints. Not only would it be able to receive input from multiple search spaces, but it would learn an architecture representation that is robust enough to accurately estimate and rank the performance of networks from unseen search spaces.

In this paper we propose a \textbf{Gen}eralized \textbf{N}eural \textbf{A}rchitecture \textbf{P}erformance \textbf{E}stimator (GENNAPE), to perform a feasibility check on pretraining a robust and generalizable neural predictor based only on open neural network benchmarks and transferring it to a wide range of unseen architecture families for accurate accuracy ranking.  GENNAPE accepts input architectures from any search space through a generic Computation Graph representation, and performs a two-stage process of graph encoding and cluster-based prediction to achieve transferability to a new neural network design space under no or minimum fine-tuning. In designing GENNAPE, we make the following contributions:

First, we propose a generic embedding scheme based on \textit{Computation Graph} (CG) representations of any neural networks and a self-supervised Contrastive Learning (CL) loss as the starting point of our predictor. Specifically, we represent an input neural network as a graph of atomic neural operations, e.g., convolutions, pooling, activation functions, rather than macro units like MBConv blocks~\cite{howard2019searching} that are specific to architecture families. This allows us to feed any neural network into our predictor. We then pass the CG through an encoder to obtain a fixed-length embedding. The graph encoder is trained by a carefully designed CL loss based on Laplacian Eigenvalues and dropout data augmentation, to distinguish graphical features between architecture families. 

\begin{table}[t]
    \centering
    \scalebox{0.82}{
    \begin{tabular}{l|c|c|c|c|c} \toprule
    \textbf{Family} & \textbf{Synflow} & \textbf{Jacov} & \textbf{Fisher} & \textbf{Snip} & \textbf{FLOPs}  \\ \midrule
    NB-201 & 0.823 & \textbf{0.859} & 0.687 & 0.718 & 0.001 \\
    NB-301 & -0.210 & -0.190 & -0.305 & -0.336 & \textbf{0.578} \\ 
    PN & 0.086 & -0.022 & 0.677 & \textbf{0.731} & 0.689 \\ 
    OFA-MBv3 & 0.648 & 0.035 & 0.602 & \textbf{0.649} & 0.614 \\
    OFA-RN & 0.724 & -0.094 & 0.651 & 0.752 & \textbf{0.785} \\ \midrule
    HiAML & 0.154 & -0.010 & -0.102 & -0.168 & \textbf{0.277} \\
    Inception & 0.085 & 0.177 & -0.152 & -0.061 & \textbf{0.412} \\
    Two-Path & 0.227 & -0.071 & 0.042 & -0.027 & \textbf{0.333} \\\bottomrule
    \end{tabular}
    }
    \caption{Spearman Rank Correlation coefficients for four Zero-Cost Proxy methods and a simple FLOPs-based predictor on all target families. Best results in bold.}
    \label{tab:zcp}
    %\vspace{-4mm}
\end{table}

Second, we propose to learn multiple prediction heads, each being an MLP and focusing on learning the characteristics of a local cluster of architectures in the training family. We use Fuzzy C-Means (FCM) clustering to partition the training family (in our case, NAS-Bench-101 due to its abundance of labelled samples) into soft clusters, where a given input architecture can find soft memberships among the clusters. Based on the FCM partitions, we construct a Multi-Layer Perceptron-Ensemble (MLP-E) that 
makes prediction for an architecture by soft aggregating the MLP heads according to the architecture's membership.  

We verify the performance of GENNAPE on a wide range of public NAS families. Specifically, we pretrain our predictor on NAS-Bench-101, and treat the search spaces of NAS-Bench-201, NAS-Bench-301, ProxylessNAS~\cite{cai2018proxylessnas}, OFA-MobileNetV3, OFA-ResNet as unseen target families. We show that our predictor is able to achieve over 0.85 SRCC and over 0.9 NDCG on all of these target families with minimal fine-tuning. When integrated into a downstream NAS search algorithm, GENNAPE further improves high-accuracy architectures from NAS-Bench-101, NAS-Bench-201 and OFA-ResNet on ImageNet120. 

To stress test the generalizability of GENNAPE, we further introduce three new challenging architecture benchmark sets: HiAML, Inception and Two-Path. These families contain certain properties, such as narrow accuracy range and ties, which make it difficult for a predictor to accurately learn architecture rankings, and thus serve as challenging benchmarks for future NAS research. We show that GENNAPE can obtain over 0.78 NDCG on all three families. Finally, we open-source\footnote{https://github.com/Ascend-Research/GENNAPE} these new benchmarks to facilitate further research on generalizable neural predictors.

%% file: src/related.tex
\section{Related Work}
\label{sec:related}

Neural predictors rely on benchmark datasets, namely NAS-Bench-101~\cite{ying2019nasbench101}, 201~\cite{dong2020nasbench201} and 301~\cite{zela2022surrogate}, which allow for performance \textit{querying}. The downside to these methods is that every architecture must be trained from scratch, possibly multiple times. Arguably, supernet-based methods like OFA~\cite{cai2020once} fall under a unique category of benchmark, where sampled architectures can be immediately evaluated. However, this is still less resource intensive than training the network from scratch. Therefore, while benchmarks provide a rich repository of architecture statistics, an upfront resource cost is incurred, motivating the need for generalized predictors that can perform well on unseen families.

One method of low-resource neural prediction are Zero-Cost Proxies (ZCP); \citet{abdelfattah2021zero} outline many variants. ZCP schemes involve gradient calculations as well as parameter salience. As Table~\ref{tab:zcp} shows, while ZCP methods may achieve good results on some architecture families, their performance is inconsistent across families. In addition, ZCP methods require network instantiation 
to perform forward passes to compute corresponding measures, while neural predictors do not.

Contrastive Learning (CL) is based on distinguishing pairs of similar and dissimilar objects. This approach has been recently used successfully in image classification~\cite{chen2020simple}. Specifically, CL classifiers improve transferability on unseen data. Furthermore, fine-tuning on a very small fraction of the labeled data from a new dataset is enough to produce state of the art classifiers. \citet{khosla2020supervised} further extend this approach to the case when image labels are known during training. We focus on applying CL methods for finding vector representations of graphs.

Some deep learning schemes use Fuzzy C-Means clustering (FCM) to design interpretable models. \citet{yeganejou2020fuzzy} substitute the last MLP for an FCM clustering layer and classify images according to the highest membership value. By contrast, GENNAPE incorporates FCM into an MLP ensemble, using the membership values to perform a weighted sum of MLP heads. 

%% file: src/family.tex
\section{Architecture Families}
\label{sec:fam}

\begin{table}[t]
    \centering
    \scalebox{0.8}{
    \begin{threeparttable}[hb]
    \begin{tabular}{l|l|c|c|c}
    \toprule
    Family & Dataset & $\#$CGs & Acc. Dist [\%] & Range [\%] \\ \midrule
    NB-101 & CIFAR-10 & 50k & 89.67 $\pm$ 5.73 & [10.00, 94.09] \\
    NB-201 & CIFAR-10 & 4.1k\tnote{$\dagger$} & 89.44 $\pm$ 6.13 & [46.25, 93.37] \\
    NB-301 & CIFAR-10 & 10k & 93.03 $\pm$ 0.74 & [88.28, 94.67] \\
    PN & ImageNet & 8.2k & 75.41 $\pm$ 0.09 & [71.15, 77.81] \\
    OFA-MBv3 & ImageNet & 7.5k & 76.94 $\pm$ 0.08 & [73.56, 78.83] \\
    OFA-RN & ImageNet & 10k & 78.19 $\pm$ 0.07 & [75.25, 79.94] \\ \midrule
    HiAML & CIFAR-10 & 4.6k & 92.36 $\pm$ 0.37 & [91.11, 93.44] \\
    Inception & CIFAR-10 & 580 & 91.99 $\pm$ 0.75 & [89.08, 94.03] \\ 
    Two-Path & CIFAR-10 & 6.9k & 89.94 $\pm$ 0.83 & [85.53, 92.34] \\\bottomrule
    \end{tabular}
    \begin{tablenotes}\footnotesize
	    \item{$\dagger$}We only use the 4096 architectures that do not contain the `none' operation.
	\end{tablenotes}
	\end{threeparttable}
	}
	\caption{Architecture families 
    in terms of the number of architecture CGs, dataset and accuracy distribution in terms of mean and standard deviation (Acc. Dist) as well as range.  Horizontal line demarcates public families.}
    \label{table:datasets}
	%\vspace{-3mm}
\end{table}

\begin{figure}[t]
    \centering
    \includegraphics[width=3.2in]{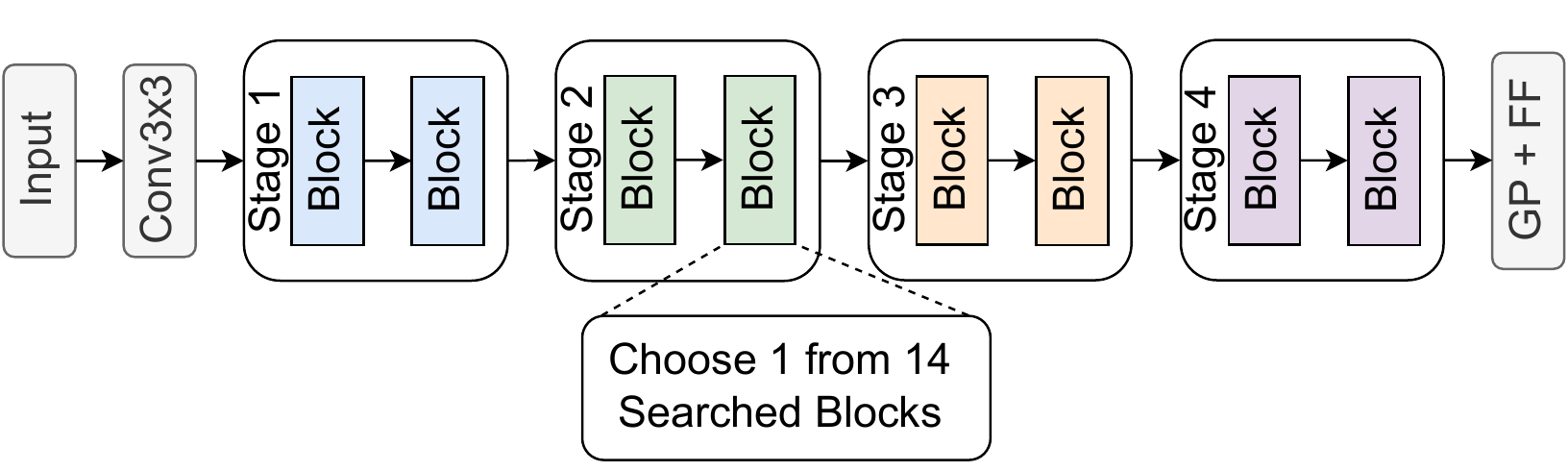}
    \caption{The architecture backbone of HiAML, containing 4 stages. Each stage contains 2 identical blocks.}
    \label{fig:hiaml}
    %\vspace{-3mm}
\end{figure}

We describe the neural network families used in this paper. We start with publicly available benchmarks before introducing three new families of networks. Table~\ref{table:datasets} provides an overall summary of each family. 

\subsection{Public NAS Benchmark Families}
\label{sec:pubData}

\textbf{NAS-Bench-101} (NB-101) is one of the first and largest benchmarks for NAS. It consists of 423k unique architectures, individually evaluated on CIFAR-10. The architectures are cell-based, where each cell is a Directed Acyclic Graph (DAG) containing operations, stacked repeatedly to form a network. We sample 50k random architectures from this family to form our CG training family. 

\textbf{NAS-Bench-201} (NB-201) and \textbf{NAS-Bench-301} (NB-301) are two additional benchmarks. Like NB-101, architectures consist of a fixed topology of cells, except they follow the DARTS search space. Additionally, NB-201 only contains 15.6k evaluated architectures, while NB-301 is a surrogate predictor for the DARTS search space. Therefore, we treat both as test families. 

\textbf{ProxylessNAS} (PN)~\cite{cai2018proxylessnas} and \textbf{Once-for-All-MobileNetV3} (OFA-MBv3)~\cite{cai2020once} are based on the MobileNet~\cite{howard2019searching} architecture families, with PN and OFA-MBv3 implementing versions 2 and 3, respectively. \textbf{Once-for-All-ResNet} (OFA-RN) is based on the classical ResNet-50~\cite{he2016deep} topology. All three evaluate on ImageNet. Architectures consist of searchable, macro features where the number of blocks is variable. We refer the reader to \citet{mills2021profiling} for further details regarding %the 
PN, OFA-MBv3 and OFA-RN. % families.

\begin{figure}[t]
    \centering
    \subfloat[HiAML Accuracy]{\includegraphics[width=1.45in]{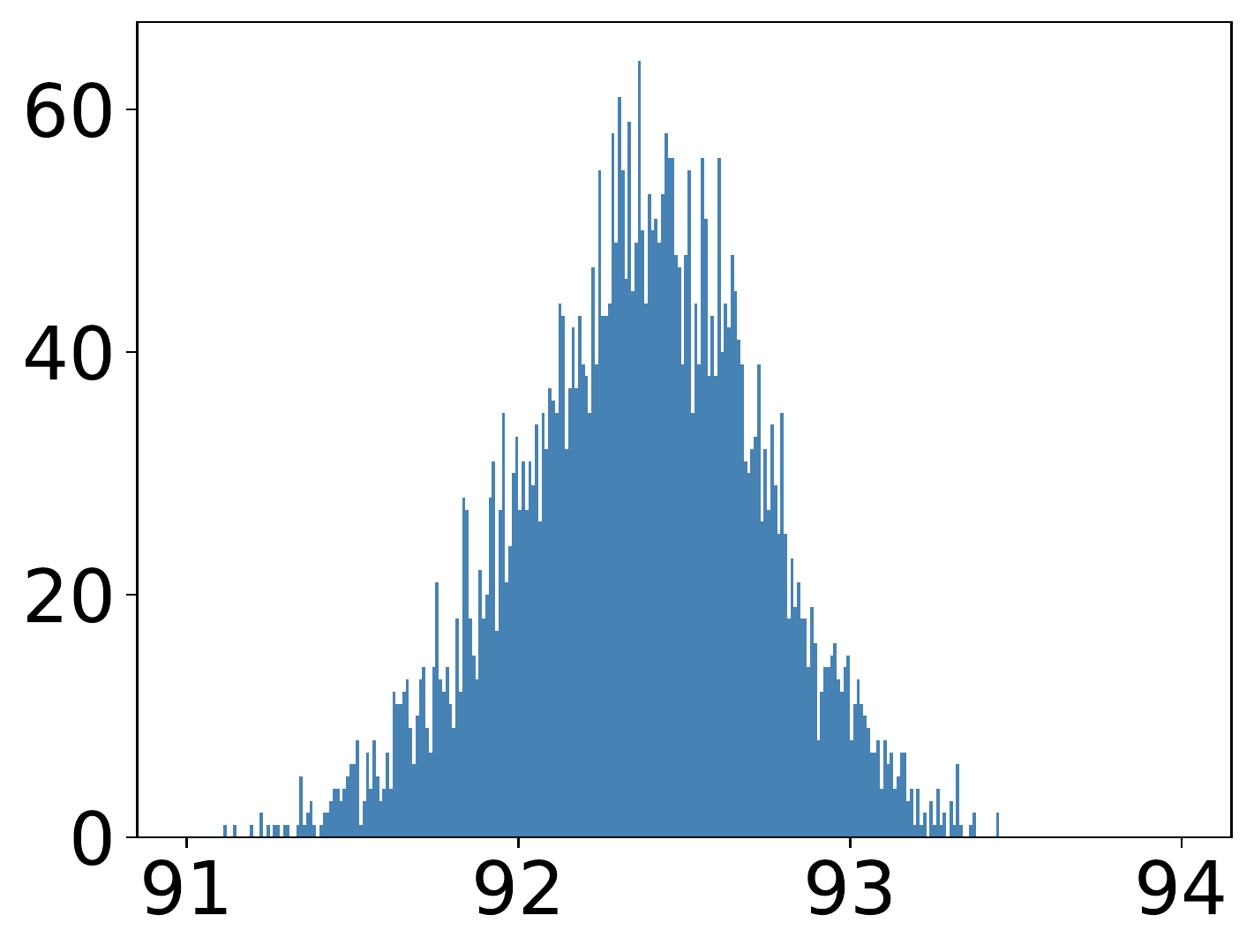}}
    \qquad
    \subfloat[HiAML FLOPs]{\includegraphics[width=1.45in]{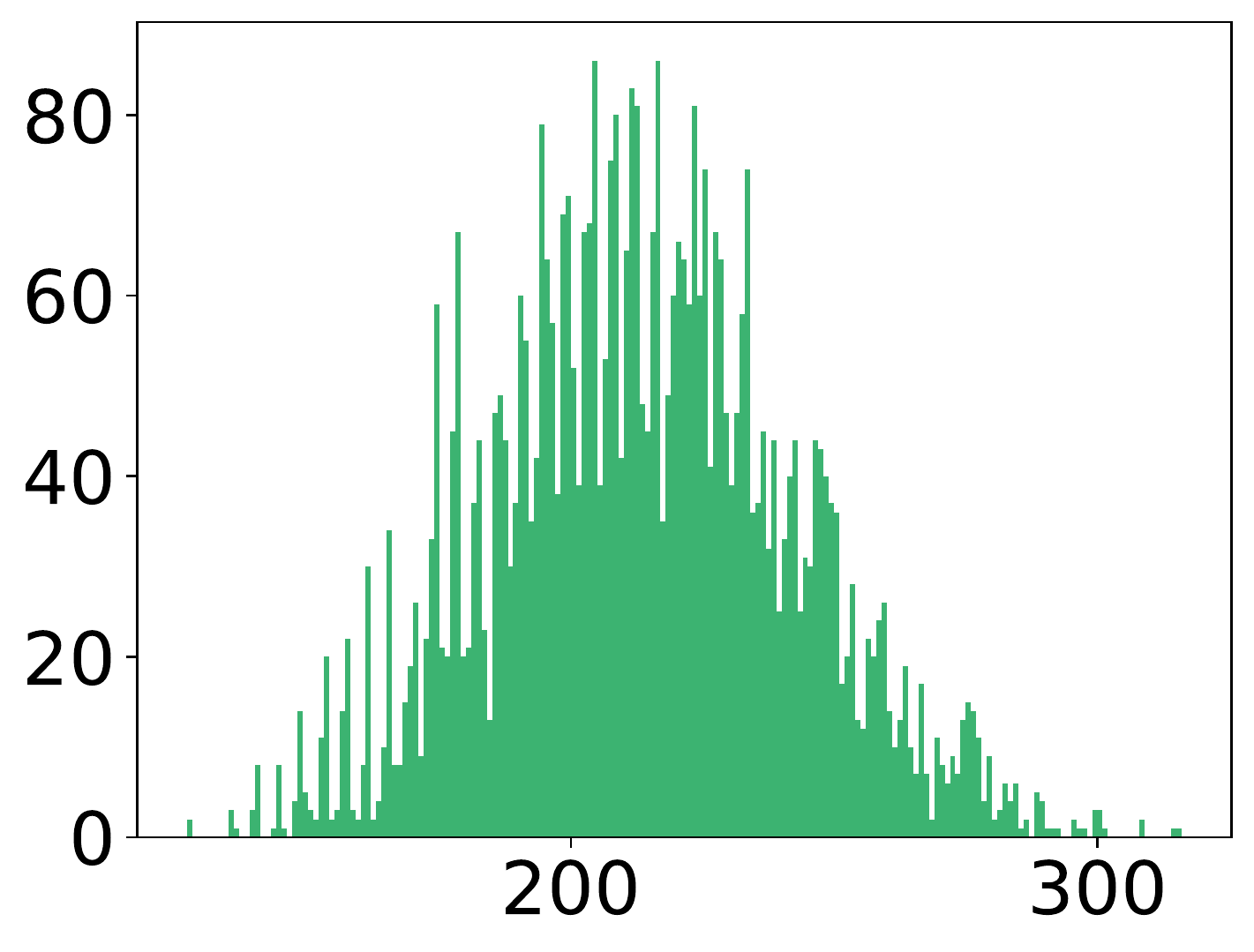}}
    \qquad
    \subfloat[Inception Accuracy]{\includegraphics[width=1.45in]{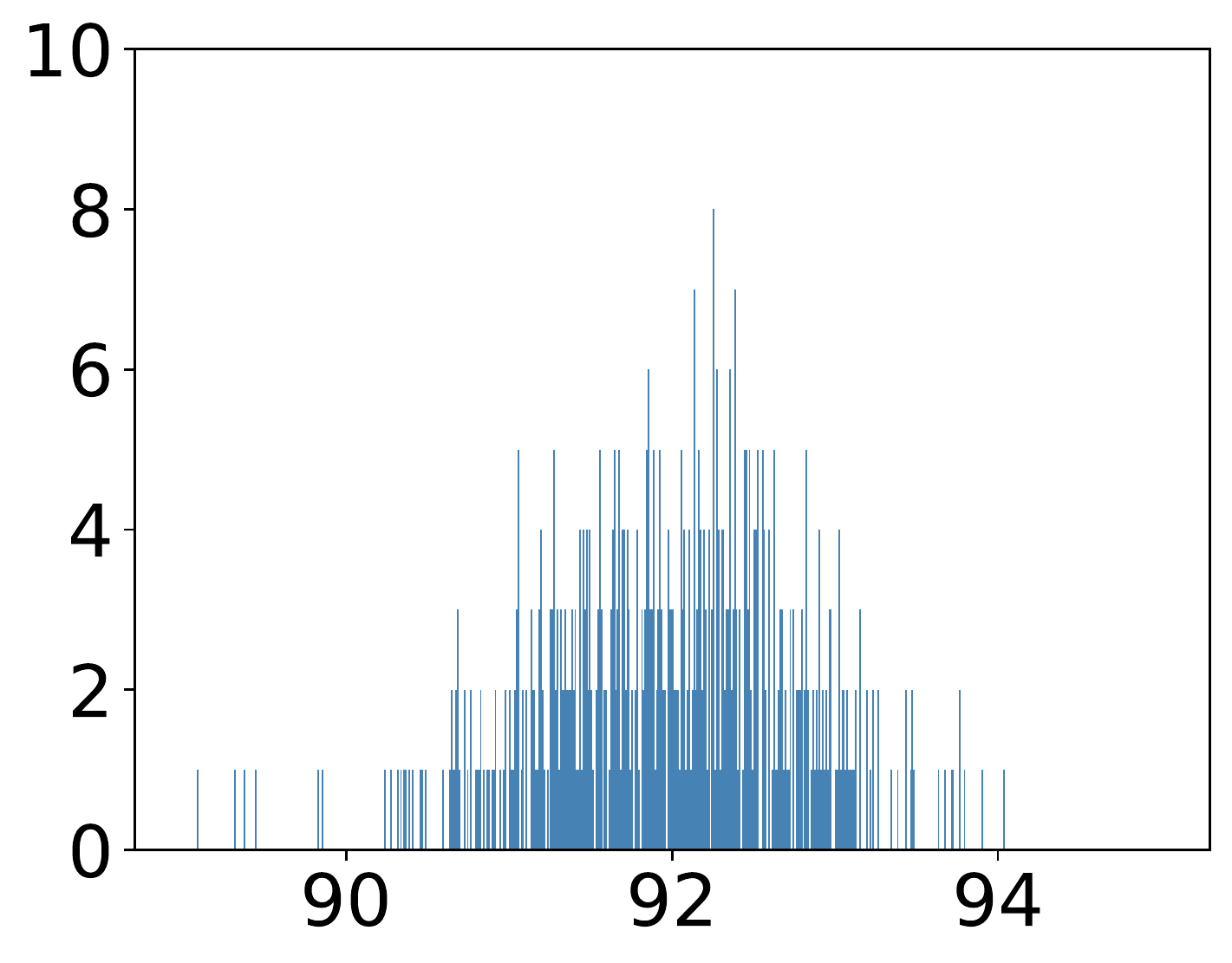}}
    \qquad
    \subfloat[Inception FLOPs]{\includegraphics[width=1.45in]{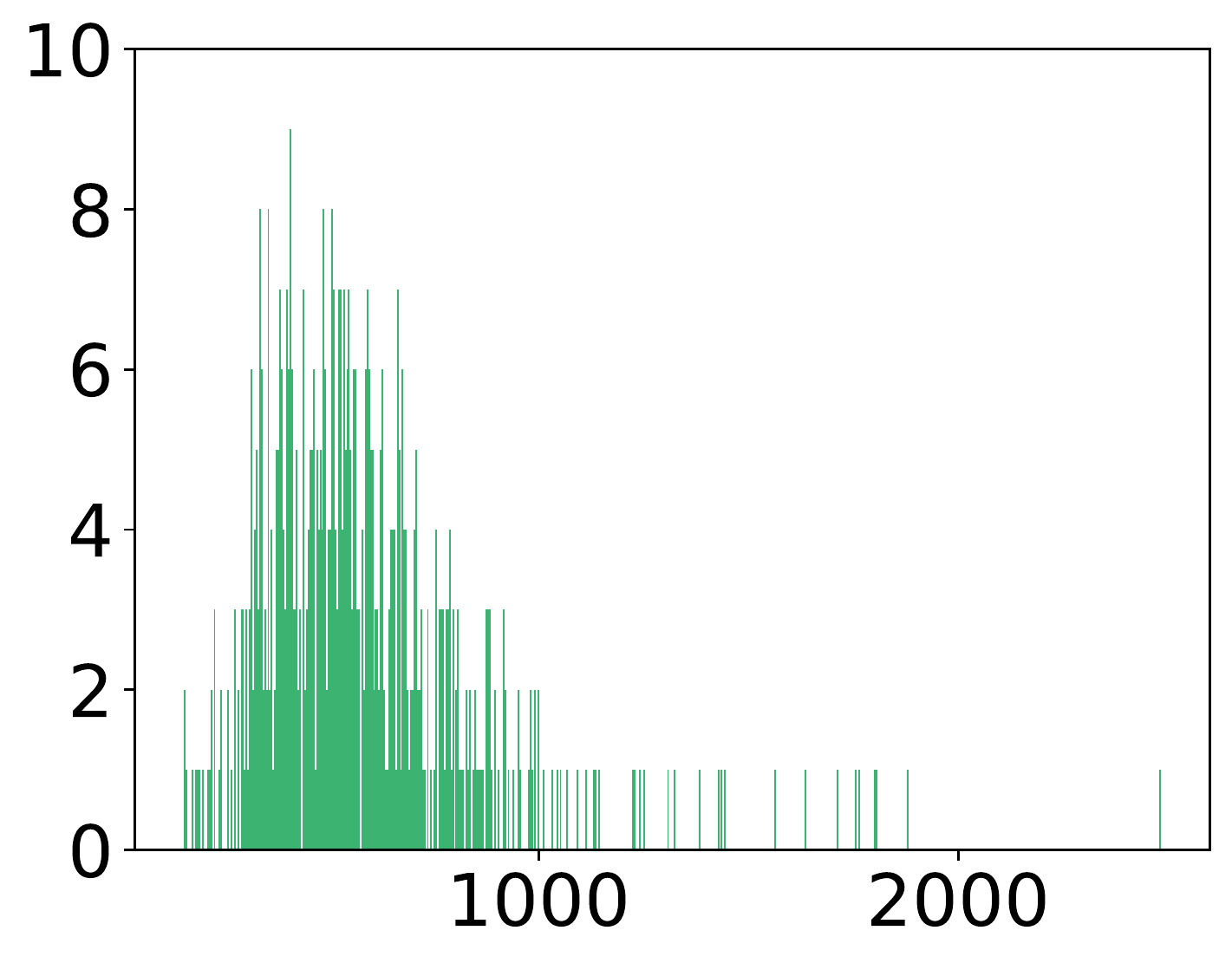}}
    \qquad
    \subfloat[Two-Path Accuracy]{\includegraphics[width=1.45in]{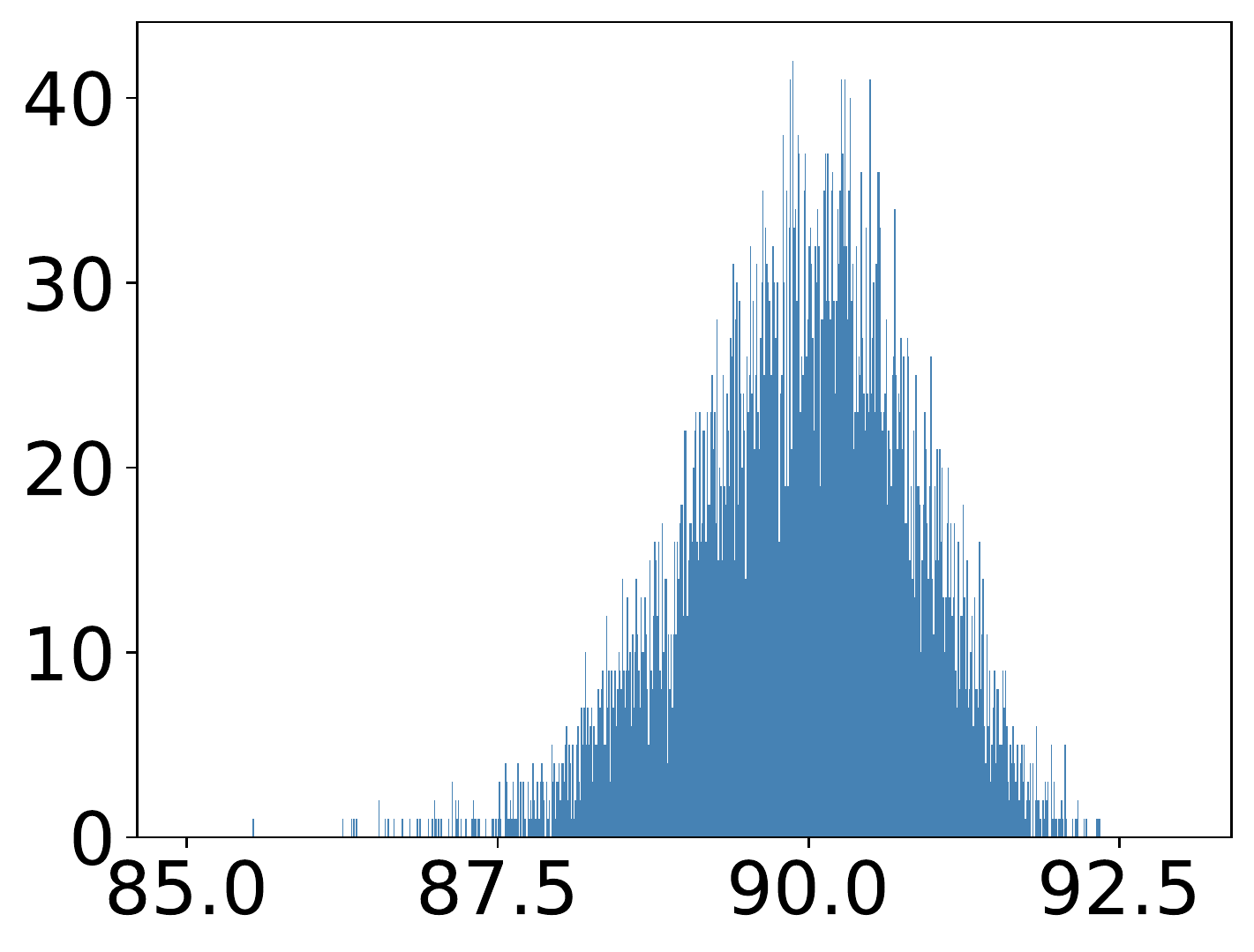}}
    \qquad
    \subfloat[Two-Path FLOPs]{\includegraphics[width=1.45in]{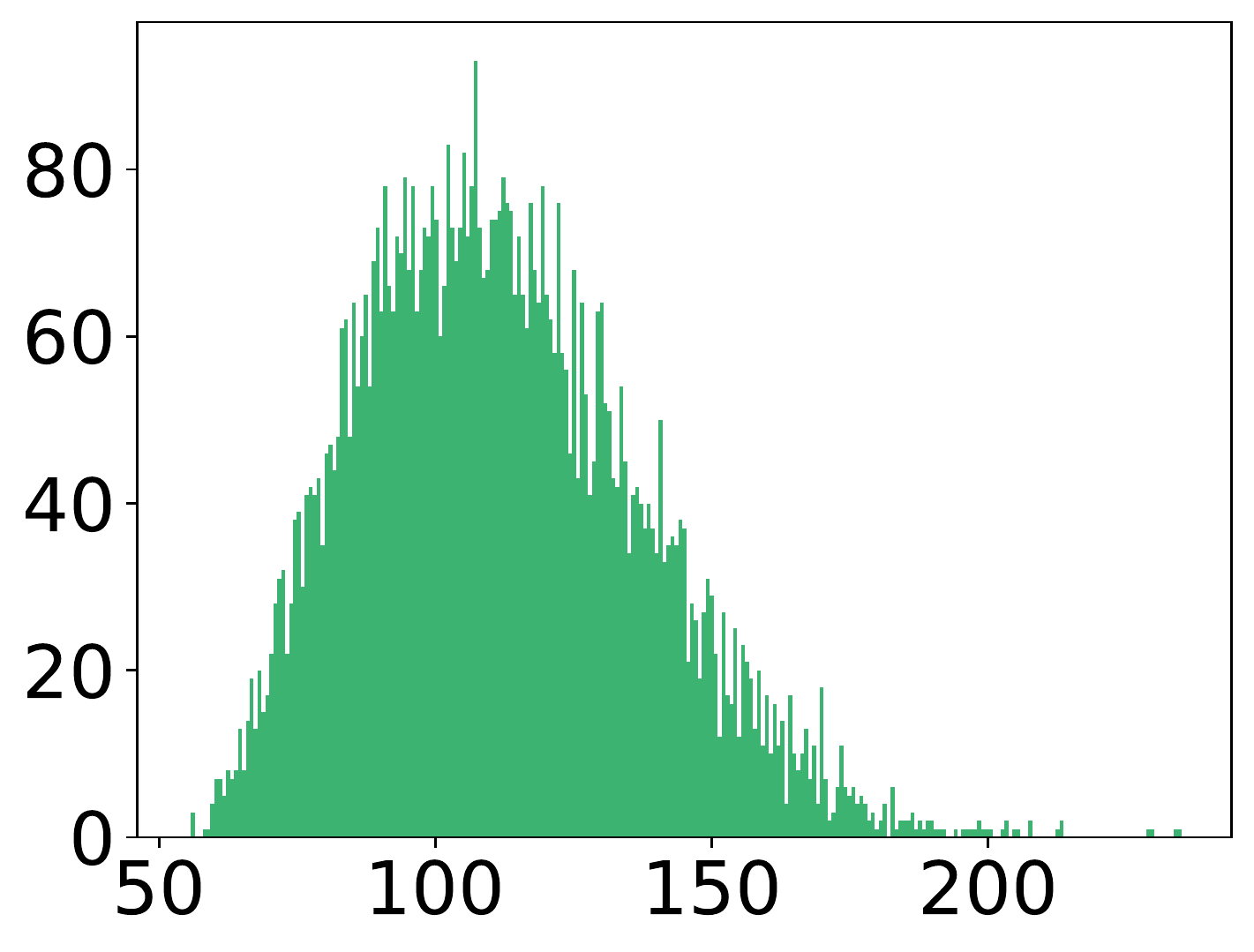}}
    \caption{Accuracy [\%] and FLOPs [10e6] histograms for the new benchmarks. Accuracy binning interval is 0.01\%.}
    \label{fig:all_hists}
\end{figure}

\subsection{Introduced NAS Benchmark Families}
\label{sec:newData}

We introduce the key attributes of the new network families and provide additional information in the supplementary. For all families, we define an \textit{operator} as a bundle of primitive operations, e.g.,  Conv3x3-BN-ReLU. A \textit{block} is a set of operators with different connection topologies, while a \textit{stage} contains a repetition of the same block type and the \textit{backbone} determines how we connect blocks to form distinct networks. For these network families, accuracy values are rounded to four decimal places. That is, we would represent 91.23\% as a terminating 0.9123. Figure~\ref{fig:all_hists} provides accuracy and FLOPs histograms. 

\begin{figure*}[t]
    \centering
    \includegraphics[width=6.9in]{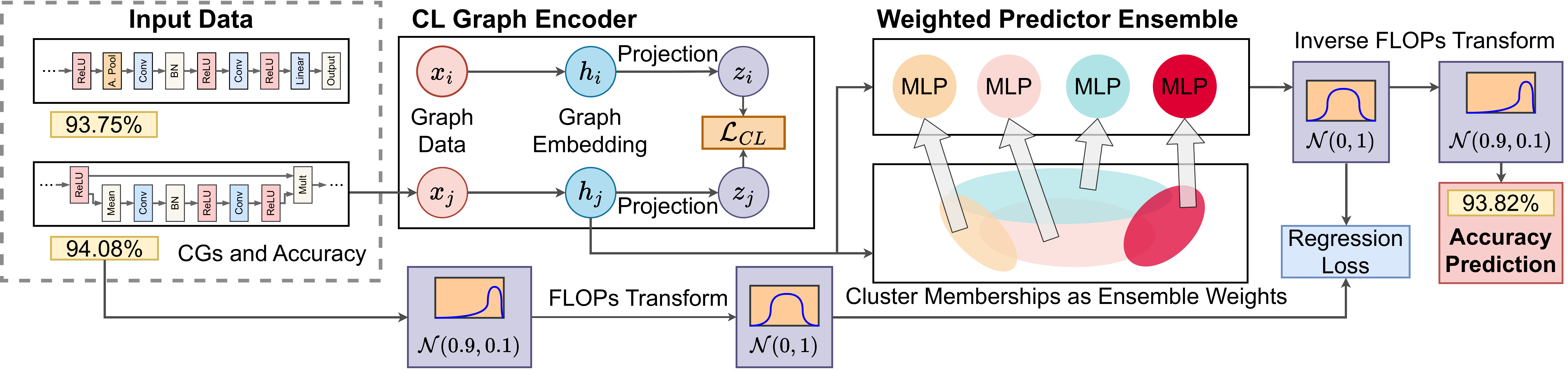}
    \caption{An overview of GENNAPE. We utilize Computation Graphs and a Contrastive Learning Graph Encoder to generate embeddings, which are fed into a Fuzzy C-Means clustering algorithm. Cluster memberships act as weights in an MLP ensemble. Finally, we use FLOPs to apply a linear transform to the accuracy.}
    \label{fig:gennape}
\end{figure*}

\textbf{HiAML} is a custom family of CIFAR-10 networks inspired by NAS-Bench-101. Networks are built using 14 pre-defined operation block structures found by GA-NAS~\cite{rezaei2021generative}. Specifically, a pair of identical blocks form a stage, and there are 4 stages per network. Since August 2021, HiAML networks have been in use as the backbone feature extractor in Huawei's mobile Facial Landmark Detection application, i.e., the family is not limited to Image Classification. HiAML networks are efficient enough to run entirely on the less powerful Ascend Tiny Core, which is a part of Da Vinci Architecture 2.0 in Huawei's Neural Processing Unit (NPU). Figure~\ref{fig:hiaml} provides a backbone illustration of HiAML.
As shown in Table~\ref{table:datasets}, we observe that 
HiAML has the narrowest performance profile among all families considered, as 
4.6k architectures fit into an accuracy range of 
only 2.33\%. As a comparison, OFA-MBv3 has the second smallest accuracy range of 4.25\%. 
This characteristic makes accurate prediction on  
HiAML networks very difficult as there are many ties as Figure~\ref{fig:all_hists}(a) shows.

\textbf{Inception} takes inspiration from the Inception-v4~\cite{szegedy2016rethinking} classification networks in literature. Our custom Inception networks consist of a single-path backbone with 3 stacked stages, and inside each stage, there is a repetition of 2 to 4 blocks. Like the original Inception, each block may contain up to 4 branching paths. The main advantage of Inception is that channels are divided among the paths, leading to lower computation costs. The backbone follows a single-path topology with 3 stages. Currently, Inception networks serve as feature extractors for Huawei's Facial Recognition framework.

\textbf{Two-Path} networks are designed to be more lightweight than other families. Like the other benchmarks, Huawei utilizes Two-Path networks for real-world applications, specifically multi-frame Super Resolution and 4k LivePhoto applications. As the name suggests, the backbone consists of two branching paths. On each path, there are 2 to 4 blocks, each a sequence of up to 3 operators. One may view the Two-path family as the complement of the Inception family. An Inception network has a single-path backbone with branching paths inside blocks. But a Two-Path network has two branching paths of blocks, while each block is a single path of operators.

%% file: src/method.tex
\section{Methodology}
\label{sec:method}

Figure~\ref{fig:gennape} visualizes our scheme. A predictor that generalizes to arbitrary architecture families should accept a generalizable neural network representation as input. We consider the Computation Graph (CG) of a network structure based on the graph structures deep learning libraries like TensorFlow~\cite{tensorflow2015-whitepaper} and PyTorch~\cite{NEURIPS2019_9015} use to facilitate backpropagation. Specifically, nodes refer to atomic operators in a network, e.g., Convolution, Linear layers, Pooling layers, non-linearities and tensor operations like mean, add and concat. Edges are directed and featureless. 

This approach allows us to encode architectures from different families using a common format that represents all primitive network operations and the connections between them. For example, both NB-101 and NB-201 include `Conv3x3' and `Conv1x1' as candidate operations. These candidates are actually sequences of three primitive operations: Convolution, ReLU and Batch Normalization (BN), which are arranged differently for each search space. Wheras NB-101 consider a `Conv-BN-ReLU' ordering, NB-201 use `ReLU-Conv-BN'. Moreover, NB-101 encodes operations as nodes, while NB-201 follow DARTS and encode operations on edges. Our CG format reconciles these differences by representing each primitive as its own node with edges defined by the actual forward-pass of the network.

\begin{figure}[t]
    \centering
    \includegraphics[width=0.83in, angle=90]{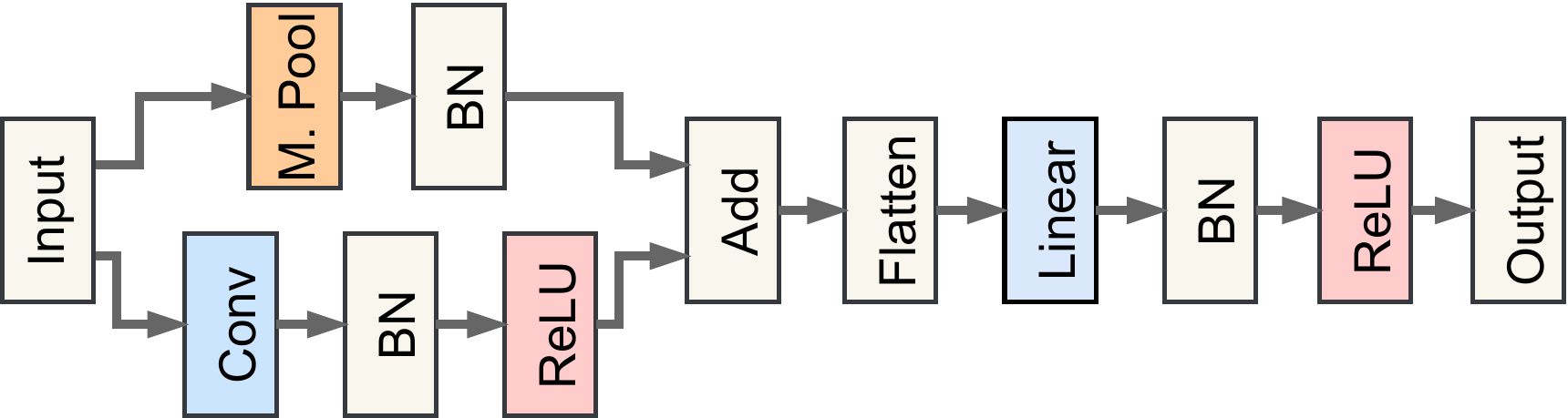}
    \caption{Example of a Compute Graph. Two parallel branches performing convolution and pooling operations process the input. A linear layer and activation then produce the output.} 
    \label{fig:simpleCG}
\end{figure}

Figure~\ref{fig:simpleCG} illustrates a simple CG, where each node is a primitive operation. Node features consist of a one-hot category for operation type as well as the height-width-channel (HWC) size of the input and output tensors. If an operation contains trainable weights, e.g., convolution and linear layers, we include a feature for the weight tensor dimensions, and a boolean for whether bias is enabled.

As graphs can contain an arbitrary number of nodes and edges, it is necessary to process them into a common, fixed-length format prior to prediction. Therefore, in this section, we introduce our Contrastive Loss-based (CL) graph embeddings scheme, the Fuzzy C-Means (FCM) ensemble that forms the GENNAPE predictor and describe a FLOPs-based accuracy transform. 

\subsection{Embeddings from Contrastive Learning}
\label{sec:cl}

We use Contrastive Learning (CL) to learn vector representations of CGs by adapting the two key components of SimCLR~\cite{chen2020simple}. The first is data augmentation, where perturbations produce two different copies of every image in a batch. The second is the CL loss function, which aims to identify positive and negative pairs within a batch. Unlike images, random perturbations of CGs are unlikely to belong to the same architecture family. Therefore, to generate embedding variation, we adopt dropout augmentation from~\citet{gao2021simcse}. 

\begin{figure}[t]
    \centering
    \includegraphics[width=3.2in]{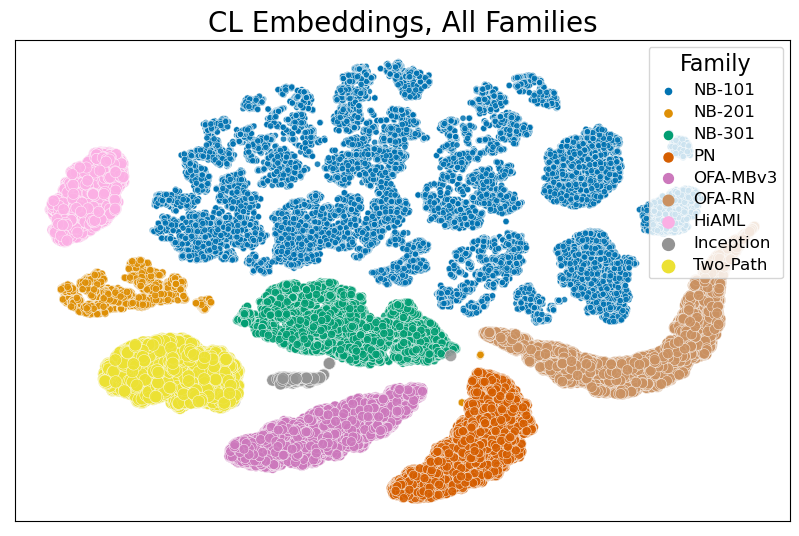}
    \caption{t-SNE scatterplot of the Contrastive Learning embeddings for all architecture families. Best viewed in color.}
    \label{fig:tsne}
\end{figure}

Consider a batch of $N$ CGs. We use dropout and apply a graph encoder twice to make $2$ copies of each CG. We then use an MLP head to project these embeddings of length $e$ into a lower dimensional space $p$ to optimize the CL loss. Let $I=\{h_1,h_2,\dots,h_{2N}\}\subset\mathbb{R}^e$ be the vector representations, and let $z_i=proj(h_i)\in \{||z||=1: z \in \mathbb{R}^p\}$ be their projections. We define the cosine similarity $sim(z_i,z_j)=z_i\cdot z_j/\tau$, with temperature $\tau>0$ and $\cdot$ as the dot product. Then, our CL loss function is
\begin{equation}
    \centering
    \label{eq:cl_loss}
    \mathcal{L}_{CL} = - \sum_{i\in I} \sum_{\ell \not=i} \alpha_\ell^{(i)} \log
    \frac{\exp(sim(z_i,z_\ell))}{\sum_{r\not=i} \exp(sim(z_i,z_r))} ,
\end{equation}
where $\alpha_\ell^{(i)} \ge 0$ and $\sum_{\ell \not= i} \alpha_\ell^{(i)}=1$. For each computational graph $i$, we calculate the convex combination $\alpha_*^{(i)}$ using the spectral distance of the underlying undirected graph as a similarity measure. We use topological properties to group architectures in the same family or separate architectures from different families, because the arrangement of operations differs for each family. Therefore, given a computation graph, let $G$ be its underlying undirected graph. The normalized Laplacian of $G$ encodes important connectivity features. Furthermore, it allows us to define a (pseudo) distance between computation graphs by evaluating the Euclidean distance between the corresponding $q = 21$ smallest eigenvalues~\cite{dwivedi2021generalization}. We use $\sigma(g_1,g_2)$ to denote the distance between computation graphs $g_1,g_2$ and define $\alpha_*^{(i)}$ to be the softmax of $\sigma(i,*)$. 

We train a CL encoder on NB-101 and use it to infer embeddings for all test families. A 2-dimensional representation of the embeddings is shown in Figure~\ref{fig:tsne}. We note two advantages of the CL encoder. First, it effectively separates each test family into distinct clusters with minimal overlap. By contrast, the CL encoder divides the training family, NB-101, into many smaller clusters. This is an important finding. While NB-101 contains many architectures, it only contains 3 candidate operations, meaning that most of the diversity amongst these architectures comes from topological differences, i.e., Laplacian eigenvalues, which play an important role in training our CL encoder. As such, each NB-101 cluster may reside closer to a given test family than others. Next, we exploit these observations to build a predictor ensemble.

\subsection{Soft-Clustering Ensemble}
\label{sec:fcm}

We use MLP Ensembles (MLP-E) in GENNAPE to make predictions. Each head in the MLP-E is is responsible for a region of the embedding space, although these regions can partially overlap with each other. That is because we use Fuzzy C-Means (FCM)~\cite{bezdek1984fcm} clustering to divide up the embedding space, and FCM represents data using continuous \textit{memberships} with respect to all clusters. The output of the MLP-E is defined as,

\begin{equation*}
    \centering
    \label{eq:mlp_ensemble}
    y'_i = \sum_{j=1}^{C}U_{i,j}f^{j}(h_i),
\end{equation*}
where $h_i$ and $y'_i$ are the graph embedding and target prediction a CG, of which there are $N$ in total, $C$ is the number of heads, $f^j$ is a head and $U_{i, j}$ is the membership for CG $i$ across cluster $j$. FCM membership values $U \in (0, 1)^{N \times C}$ are positive numbers that sum to 1 for a given sample; $\forall i \sum_{j=1}^{C} U_{i, j} = 1$.

Like other distance-based clustering algorithms, FCM consists of centroids that are computed alongside membership values by minimizing an optimization. Let $d_{i,j}$ be the Euclidean distance between CG embedding $h_i$ and FCM centroid $v_j$. FCM converges by alternating between updating membership values and centroid locations, respectively, according to the following equations,

\begin{equation}
    \centering
    \label{eq:fcmMembership}
    U_{i, k} = (\sum_{j=1}^{C}(\dfrac{d_{i,k}^2}{d_{i,j}^2})^{\dfrac{1}{m-1}})^{-1},
\end{equation}

\begin{equation}
    \centering
    \label{eq:fcmCentroids}
    v_k = \dfrac{\sum_{i=1}^N U_{i,k}^m h_i}{\sum_{i=1}^N U_{i,k}^m},
\end{equation}
where $m > 1$ is known as the \textit{fuzzification coefficient}, and is typically a value of 2 or greater~\cite{pedrycz2012genetic}, which controls the degree of overlap between the clusters. 

The FCM algorithm alternates between Equations~\ref{eq:fcmMembership} and \ref{eq:fcmCentroids} until either a pre-specified number of iterations has been reached, or the following stopping criteria is satisfied: 

\begin{equation}
    \centering
    \label{eq:fcmStop}
    \max_{i, j} \lvert U_{i, j}(t+1) - U_{i,j}(t) \rvert \leq \epsilon, 
\end{equation}
where $t$ is a time parameterization of $U$. Essentially, clustering convergence is reached once the maximum change in membership across all $N$ data samples and $C$ clusters is less than $\epsilon = 1e^{-9}$. 

When applying FCM to GENNAPE, we compute cluster centroids on the training family. For test families, we infer membership from existing centroids using Equation~\ref{eq:fcmMembership}. 

A simple MLP is not guaranteed to generalize to features of families it has not seen. However, we can aid generalizability by only having the MLP learn a specific region in the feature space and weighing its predictions based on the distance between the unseen features and the learnt region. Moreover, we can divide up the feature space with an MLP-E. Each head learns to specialize on a subset of the embedding features for the training family, as well an association with nearby, unseen features, depending on memberships. 

\subsection{FLOPs-based Features and Transforms}
\label{sec:FLOPs}

Every network has an associated FLOPs value, or floating point operations required for the forward pass. GENNAPE uses FLOPs to transform the regression targets. While the downstream goal of any regressor is to estimate the accuracy of a neural network, GENNAPE estimates the following, 

\begin{equation}
    \centering
    \label{eq:accTrans}
    y_i = \mathcal{Z}(\dfrac{A_i}{\texttt{Log}_{10}(F_i+1)+1}),
\end{equation}
where $y_i$ is the label, $A_i$ is the accuracy of a network, $F_i$ is the FLOPs, measured in gigaFLOPs and $\mathcal{Z}$ is the Z-Score. The inverse of this transformation can be applied to a prediction to obtain the corresponding accuracy prediction. We use the standardization statistics from the NB-101 training set for all test families, and we prune out NB-101 architectures whose accuracy is below 80\% when implementing Equation~\ref{eq:accTrans}. This leaves us with 49k architectures.

%% file: src/results.tex
\section{Experimental Results}
\label{sec:exp}

In this Section, we describe our experimental procedure and results in terms of regression and correlation metrics. Finally, we apply GENNAPE to NAS and perform search. 

\subsection{NAS-Bench-101 Comparison}
\label{sec:setup}

We use NB-101 as the training family and treat all others as unseen test families. We split the NB-101 architectures into a training set with 80\% of the data, and two separate validations sets that contain 10\% each. We apply Equation~\ref{eq:accTrans} using the mean and standard deviation from the training partition to standardize the validation sets. The model trains for 40 epochs. We evaluate it on the first validation set at every epoch without applying the inverse of Equation~\ref{eq:accTrans} to track loss statistics on transformed labels. Once training is complete, we further evaluate the model on the second validation set, this time applying the inverse of Equation~\ref{eq:accTrans} on predictions to calculate performance compared to the ground truth accuracy. For each scheme, we train a model 5 times using different random seeds. The overall best model is the one that achieves the highest rank correlation on the second NB-101 validation set. We provide hyperparameters and other training details in the supplementary materials. 

\begin{table}[t]
    \centering
    \scalebox{0.8}{
    \begin{threeparttable}
    \begin{tabular}{l|c|c} \toprule
    \textbf{Method} & \textbf{MAE} & \textbf{SRCC} \\ \midrule
    NPN\tnote{$\dagger$} & 1.09 $\pm$ 0.01\% & \textbf{0.934 $\pm$ 0.003} \\
    BANANAS\tnote{$\dagger$} & 1.40 $\pm$ 0.06\% &0.834 $\pm$ 0.002 \\
    TNASP\tnote{$\dagger$} & 1.23 $\pm$ 0.02\% & 0.918 $\pm$ 0.002\\ \midrule
    GCN & 1.78 $\pm$ 0.06\% & 0.732 $\pm$ 0.034 \\
    GIN & 1.72 $\pm$ 0.04\% & 0.735 $\pm$ 0.035 \\
    $k$-GNN & 1.61 $\pm$ 0.08\% & 0.814 $\pm$ 0.020 \\
    CL & 1.51 $\pm$ 0.17\% & 0.874 $\pm$ 0.009 \\
    CL+FCM & 1.19 $\pm$ 0.12\% & 0.896 $\pm$ 0.003 \\
    CL+T & \textit{0.65 $\pm$ 0.08\%} & 0.921 $\pm$ 0.003 \\
    CL+FCM+T& \textbf{0.59 $\pm$ 0.01\%} & \textit{0.930 $\pm$ 0.002} \\ \bottomrule
    \end{tabular}
    \begin{tablenotes}\footnotesize
	    \item{$\dagger$}Obtained by running their model using our 50k NB-101 models, using the same data splits.
	\end{tablenotes}
	\end{threeparttable}
    }
    \caption{NB-101 performance in terms of MAE [\%] and SRCC. We compare our CL-based schemes to various regressor-based neural predictors. Horizontal line separates non-transferable and transferable predictors. Results averaged over 5 runs.}
    \label{tab:nb101_compare}
    \vspace{-3mm}
\end{table}

We consider several types of regressor variants in our experiments. `CL+FCM+T' uses CL graph embeddings as well as the FCM MLP-E to predict performance labels transformed using Equation~\ref{eq:accTrans}. `CL+T' removes the FCM clustering ensemble and uses a single MLP while `CL+FCM' keeps the clustering ensemble while removing Equation~\ref{eq:accTrans}. Finally, `CL' drops both Equation~\ref{eq:accTrans} and the FCM ensemble to predict accuracy directly using the CL embeddings. As baselines, we consider several GNN regressors such as $k$-GNN~\cite{morris2019weisfeiler}, Graph Convolutional Networks (GCN)~\cite{welling2016semi} and Graph Isomorphism Networks (GIN)~\cite{xu2019GIN} and train them end-to-end on CGs. Finally, we also consider several related neural predictors, such as Neural Predictor for NAS (NPN)~\cite{wen2020neural}, BANANAS~\cite{white2019bananas} and TNASP~\cite{lu2021TNASP}. These predictors operate on the NB-101 family, however since they do not represent architectures using CGs, they are not transferable outside of NB-101.

We consider two predictor performance metrics. The first is Mean Absolute Error (MAE), which measures predictor accuracy with respect to ground-truth labels. We report MAE as a percentage where lower is better. The second is Spearman's Rank Correlation (SRCC) which judges a predictor's ability to rank a population of architectures. SRCC ranges from -1 to 1 and in our case higher is better as that represents positive agreement with the ground-truth.

Table~\ref{tab:nb101_compare} enumerates our results for NB-101. We note the effectiveness of Equation~\ref{eq:accTrans}, which allows us to achieve MAE scores below 1\% on average. Moreover, the inclusion of CL and FCM further improve this result to a minimum of 0.59\%. In terms of ranking correlation, CL+FCM+T achieves the second highest SRCC at 0.930. While NPN obtains a slightly higher SRCC, that is the limit of its capabilities, since NPN on NB-101 cannot perform predictions for architectures from other families. Finally, in terms of other GNN CG baselines, the $k$-GNN excels. While it does not obtain the best MAE or SRCC, it is distinctly better than both the GCN and GIN. Therefore, we continue to use it as a generalizable predictor baseline. 

\subsection{Transferability Prediction Error}
\label{sec:family_tests}

When evaluating on test families, we consider two cases. The first is zero-shot transfer performance, where we measure a given metric on all architecture CGs in a family without fine-tuning the predictor. In the second case, we sample 50 architectures at random, fine-tune for 100 epochs and then evaluate on the remaining, held-out architectures. We perform fine-tuning 5 separate times on different seeds, ensuring that we select the same 50 architectures per seed. Evaluating MAE performance on test families is applicable in the fine-tuning setting as that provides the predictor some information on the actual distribution of target labels. When fine-tuning using Equation~\ref{eq:accTrans}, we use the mean and standard deviation information from the NB-101 training partition to standardize labels for the 50 fine-tuning samples, and to recover accuracy on the remaining samples at inference.

Table~\ref{tab:mae} provides test family MAE results for the $k$-GNN baseline and several CL-based regressors. Once again, we observe how regressors that use Equation~\ref{eq:accTrans} achieve the lowest MAE on 6 of 9 families, including all ImageNet families and Two-Path. The $k$-GNN achieves good performance on NB-201, likely because it has the second largest range after NB-101. On HiAML, Inception and Two-Path, the results are generally very close, showing how difficult it can be to predict accuracy for them. 

\begin{table}[t]
    \centering
    \scalebox{0.8}{
    \begin{tabular}{l|c|c|c|c} \toprule
    \textbf{Family} & \textbf{\textit{k}-GNN} & \textbf{CL} & \textbf{CL+T} & \textbf{CL+FCM+T} \\ \midrule
    NB-101 & 1.61 & 1.51 & 0.65 & \textbf{0.59} \\ \midrule
    NB-201 & \textbf{1.80} & 2.81 & 1.93 & 2.05 \\
    NB-301 & 0.46 & 0.85 & \textbf{0.43} & 0.57 \\ \midrule
    PN & 0.56 & 0.40 & 0.35 & \textbf{0.33} \\ 
    OFA-MBv3 & 0.61 & 0.35 & 0.31 & \textbf{0.27} \\
    OFA-RN & 0.40 & 0.44 & 0.45 & \textbf{0.37} \\ \midrule
    HiAML & \textbf{0.35} & 0.45 & 0.40 & 0.39 \\
    Inception & \textbf{0.60} & 0.82 & 0.71 & 0.86 \\
    Two-Path & 0.80 & 1.11 & \textbf{0.71} & 0.72 \\ \bottomrule
    \end{tabular}
    }
    \caption{Mean Absolute Error [\%] results across all families and regressors. Lower is better. Best result in bold. Results averaged across 5 random seeds.}
    \label{tab:mae}
    \vspace{-3mm}
\end{table}

\subsection{Unseen Architecture Ranking Performance}
\label{sec:srcc_ndcg}

GENNAPE consists of a weighted average of predictions from several models. The first two are the aformentioned `CL+FCM+T' and `CL+T' regressors. We also include two pairwise classifiers, that, instead of estimating accuracy, predict which CG in a pair has higher accuracy. We label these as `CL+Pairwise+FCM' and `CL+Pairwise', differentiated by whether they have FCM and an MLP ensemble. We also include the $k$-GNN and the FLOPs predictor from Table~\ref{tab:zcp}. For zero-shot performance, GENNAPE weighs the output of all predictors equally. When fine-tuning, we calculate weights by taking the softmax of the Kendall's Tau (KT) result each individual predictor obtains on the fine-tuning samples. The supplementary materials contains details and an ablation study of these components.

Table~\ref{tab:srcc} provides SRCC results across all test families. We observe how GENNAPE achieves SRCC values above 0.85 on all public benchmarks. In fact, the only public family GENNAPE does not achive zero-shot SRCC above 0.5 on is NB-301, however, at 0.3214 this is still much higher than the $k$-GNN and any of the ZCPs from Table~\ref{tab:zcp}. Moreover, fine-tuning greatly improves the performance of GENNAPE on NB-301. Meanwhile, the $k$-GNN does achieve higher zero-shot SRCC on OFA-RN, but the improvement is less than 0.1 SRCC. On all other public benchmarks it fails to achieve over 0.5 SRCC in the zero-shot setting.

GENNAPE achieves the best zero-shot performance on HiAML, Inception and Two-Path. With fine-tuning, it manages to achieve above 0.5 SRCC on Inception and close to 0.5 SRCC on Two-Path. By contrast the $k$-GNN fails to achieve positive correlation on HiAML and Inception in the zero-shot setting, and while it achieves better HiAML SRCC with fine-tuning, once again the performance gap is very small, below 0.015 SRCC and less than the GENNAPE zero-shot SRCC of 0.4331. Finally, GENNAPE does lose some ranking performance when fine-tuning on HiAML, but that is a feature of the family: a narrow accuracy range with many ties that is hard to predict on. We provide additional results using Kendall's Tau in the supplementary materials. 

\begin{table}[t]
    \centering
    \scalebox{0.8}{
    \begin{tabular}{l|c|c} \toprule
    \textbf{Family} & \textbf{\textit{k}-GNN} & \textbf{GENNAPE} \\ \midrule
    NB-201 & 0.4930 & \textbf{0.8146} \\
    w/ FT & 0.8606  $\pm$  0.0245 & \textbf{0.9103 $\pm$ 0.0114} \\ \midrule
    NB-301 & {0.0642} & \textbf{0.3214} \\
    w/ FT & 0.8584 $\pm$ 0.0290 & \textbf{0.8825 $\pm$ 0.0134} \\\midrule
    PN & 0.0703 & \textbf{0.8213} \\
    w/ FT & 0.7559 $\pm$ 0.0621 & \textbf{0.9506 $\pm$ 0.0039} \\ \midrule
    OFA-MBv3 & 0.4345 & \textbf{0.8660} \\
    w/ FT & 0.6862 $\pm$ 0.0253 & \textbf{0.9449 $\pm$ 0.0015} \\ \midrule
    OFA-RN & \textbf{0.5721} & 0.5115 \\
    w/ FT & 0.9102 $\pm$ 0.0146 & \textbf{0.9114 $\pm$ 0.0063} \\\midrule \midrule
    HiAML & -0.1211 & \textbf{0.4331} \\
    w/ FT & \textbf{0.4300 $\pm$ 0.0507} & 0.4169 $\pm$ 0.0479 \\ \midrule
    Inception & -0.2045 & \textbf{0.4249} \\
    w/ FT & 0.3340 $\pm$ 0.0793 & \textbf{0.5524 $\pm$ 0.0166} \\ \midrule
    Two-Path & 0.1970 & \textbf{0.3413} \\
    w/ FT & 0.3694 $\pm$ 0.0406 & \textbf{0.4875 $\pm$ 0.0311} \\ \bottomrule
    \end{tabular}
    }
    \caption{Spearman Rank Correlation results across test families 
    in the zero-shot transfer and fine-tuning (w/ FT) contexts. Higher is better. Best results in bold. Fine-tuning results averaged across 5 random seeds.}
    \label{tab:srcc}
    \vspace{-3mm}
\end{table}

%SRCC measures correlation over all $N$ samples and assigns equal weight to each. However, the concerns of NAS relate to finding high-performance architectures. Therefore, for further evaluation of GENNAPE, we consider another metric that assigns disproportionate importance. Following AceNAS~\cite{zhang2021acenas}, we adopt a metric from the field of Information Retrieval (IR), where queried objects are assigned a relevance score, and it is more important to properly order objects with higher relevance. This goal is analogous to the concerns of NAS. Accuracy is relevance. 

%Specifically, we adopt Normalized Discounted Cumulative Gain (NDCG), which prioritizes the ordering of the top-$k$ architectures by a ground truth ranking. NDCG is reported in the range $[0, 1]$, where higher is better. We consider the case of $k=10$ in our experiments, which corresponds to ordering the very best architectures. Details of the NDCG calculation are in the supplementary materials. 

SRCC measures correlation over all $N$ samples and assigns equal weight to each. However, the concerns of NAS relate to finding high-performance architectures. Therefore, for further evaluation of GENNAPE, we consider another metric that assigns disproportionate importance. Following AceNAS~\cite{zhang2021acenas}, we adopt Normalized Discounted Cumulative Gain (NDCG), an Information Retrieval (IR) metric %a metric from the field of Information Retrieval (IR), 
where queried objects are assigned a relevance score. In IR, it is more important to properly order objects with higher relevance. This goal is analogous to the concerns of NAS. Accuracy is relevance. We consider the case of $k=10$ in our experiments, which corresponds to ordering the very best architectures. Details of the NDCG calculation are in the supplementary materials. 

We list our findings in Table~\ref{tab:ndcg}. GENNAPE clearly outperforms the $k$-GNN baseline in both contexts. The sole exception is OFA-R50, likely due to NB-101 architectures being partially based on ResNets in terms of operation choice. Still, in the zero-shot transfer context GENNAPE achieves over 0.65 NDCG on \textit{all} families, and over 0.9 NDCG on all public families with fine-tuning.

Moreover, GENNAPE overtakes the $k$-GNN on HiAML. Results are also more favorable on Inception and Two-Path, where GENNAPE achieves over 0.8 NDCG in both the zero-shot and fine-tuning settings. This aligns with our earlier intuition: Although the small accuracy range makes global ranking difficult, there are only a few architectures at the tail of the distribution and GENNAPE is better at determining which ones they are. We report additional NDCG results in the supplementary materials.

In sum, we show that GENNAPE is a robust neural predictor that can easily generalize to unseen architecture families. The NDCG results we show are of particular relevance to the problem of NAS, where neural predictors fill the role of performance evaluation. This is because, ultimately, the downstream performance of any search algorithm relies upon having a performance estimator capable of identifying high-performance architectures. Next, we apply GENNAPE to the problem of NAS, to demonstrate its applied capability and the flexibility of our CG architecture representation.

\begin{table}[t]
    \centering
    \scalebox{0.8}{
    \begin{tabular}{l|c|c} \toprule
    \textbf{Family} & \textbf{\textit{k}-GNN} & \textbf{GENNAPE} \\ \midrule
    NB-201 & 0.9270 & \textbf{0.9793} \\
    w/ FT & 0.9751 $\pm$ 0.0082 & \textbf{0.9855 $\pm$ 0.0030} \\ \midrule
    NB-301 & 0.5341 & \textbf{0.7885}\\
    w/ FT & 0.9723 $\pm$ 0.0134 & \textbf{0.9765 $\pm$ 0.0081} \\\midrule
    PN & 0.4426 & \textbf{0.8736} \\
    w/ FT & 0.9287 $\pm$ 0.0271 & \textbf{0.9800 $\pm$ 0.0057} \\ \midrule
    OFA-MBv3 & 0.8464 & \textbf{0.9234} \\
    w/ FT & 0.8859 $\pm$ 0.0536 & \textbf{0.9838 $\pm$ 0.0030} \\ \midrule
    OFA-RN & \textbf{0.9470} & 0.6606 \\
    w/ FT & \textbf{0.9717 $\pm$ 0.0090} & 0.9463 $\pm$ 0.0236 \\\midrule \midrule
    HiAML & 0.5088 & \textbf{0.6892} \\
    w/ FT & 0.7356 $\pm$ 0.0371 & \textbf{0.7804 $\pm$ 0.0211} \\ \midrule
    Inception & 0.6064 & \textbf{0.8150} \\
    w/ FT & 0.7310 $\pm$ 0.0423 & \textbf{0.8073 $\pm$ 0.0072} \\ \midrule
    Two-Path & 0.6339 & \textbf{0.8275} \\
    w/ FT & 0.7860 $\pm$ 0.0268 & \textbf{0.8392 $\pm$ 0.0220} \\
    \bottomrule
    \end{tabular}
    }
    \caption{Normalized Discounted Cumulative Gain (NDCG@10) across test families 
    in the zero-shot transfer and fine-tuning (w/ FT) context. Values are reported in the range $[0, 1]$ and higher is better. Best results in bold. Fine-tuning results averaged across 5 random seeds.}
    \label{tab:ndcg}
    \vspace{-2mm}
\end{table}

\subsection{Applying GENNAPE to NAS}
\label{sec:search}

We conduct NAS experiments where we directly modify the CGs of existing optimal classification networks. We aim to reduce FLOPs while improving or maintaining accuracy using a simple location search algorithm and provide details in the supplementary materials. After search, we train and evaluate the original optimal architecture and the one and found by search on classification datasets.

%Specifically, 
We apply this %search 
routine to the best CIFAR-10 architectures in NB-101 and NB-201. % on CIFAR-10. 
We also search on the best OFA-RN architecture found by \citet{mills2021profiling} using the first 120 classes of ImageNet % a subset of ImageNet, which only contains the first 120 classes %; ImageNet120, 
to reduce carbon footprint. %To reduce carbon footprints, we %also search on OFA-RN
%but 
%use a subset of ImageNet, which only contains the first 120 classes; ImageNet120. 
We use the fine-tuning CL+FCM+T predictors in these experiments. Search takes less than 3 hours. Inference on our predictors is fast. The primary bottleneck stems from the search algorithm determining whether newly found CGs are feasible as classification neural networks. 

Table~\ref{tab:search} summarizes the results, demonstrating the power of GENNAPE to improve upon existing, high-performance architectures, in terms of accuracy and FLOPs, on three public families. Notably, the OFA-ResNet architecture found by search achieves 22\% FLOPs reduction while improving accuracy by 0.4\%. Moreover, these results demonstrate the utility of our CG framework, as the atomic operation mutations result in architectures that are not exact fits to the original search spaces. Using CGs, we can modify sections like the stem or head that are typically fixed, and introduce new operation sequences. For example, the NB-201 architecture we compare to and outperform was already the best on CIFAR-10. Therefore, a typical neural predictor for NB-201 would not be able to adequately process and predict performance for architectures found by our search algorithm.

\begin{table}[t]
    \centering
    \scalebox{0.8}{
    \begin{tabular}{l|l|c|c}
    \toprule
    Model             & Dataset     & FLOPs  & Top-1 Acc.(\%)  \\
    \midrule
    NB-101-Best        & CIFAR-10    & 11.72G & 94.97 \\
    NB-101-Search      & CIFAR-10    & \textbf{9.49}G & \textbf{95.05} \\
    \midrule
    NB-201-Best        & CIFAR-10    & 313M   & 93.27           \\
    NB-201-Search      & CIFAR-10    & \textbf{283}M   & \textbf{93.62}           \\
    \midrule
    OFA-RN-Input  & ImageNet120 & 12.13G & 80.62           \\
    OFA-RN-Search & ImageNet120 & \textbf{9.46}G  & \textbf{81.08}    \\
    \bottomrule
    \end{tabular}
    }
    \caption{Evaluation results for architectures found using CL+FCM+T predictors and our CG-based search algorithm.} 
    \label{tab:search}
    \vspace{-3mm}
\end{table}

%% file: src/conclusion.tex
\section{Conclusion}
\label{sec:conclusion}

In this paper, we introduce GENNAPE, or Generalized Neural Architecture Performance Estimators, to address issues present in neural predictor design. Namely, that neural predictors typically operate within a fixed search space, and are not generalizable to unseen architecture families. GENNAPE receives Computation Graphs representing arbitrary network architectures as input, before using a Contrastive Learning encoder to generate embeddings. The embeddings pass through an MLP Ensemble, and we compute the prediction using a weighted summation according to Fuzzy C-Means clustering memberships. We extensively test GENNAPE against a number of known neural predictors on NB-101 and show that it yields high MAE and SRCC performance. Experimental results demonstrate the generalizability of our scheme in zero-shot and fine-tuning contexts in terms of SRCC and NDCG@10. When applied to search, GENNAPE can improve upon existing, high-performance NB-101, NB-201 and OFA-ResNet architectures. Finally, we introduce three new challenge families: HiAML, Inception and Two-Path, as open-source benchmarks. 

%% file: src/supplementary.tex
\section{Supplementary Material}
\label{sec:supp}

\subsection{Introduced Neural Network Benchmarks}
\label{sec:customFamilies}

Figure~\ref{fig:families} provides a detailed depiction of the three new benchmark families we propose: HiAML, Inception and Two-Path. 
We re-iterate 
the key attributes of these new benchmarks. 
Specifically, we define an \textit{operator} as a bundle of primitive operations, e.g., Conv3x3-BN-ReLU. A \textit{block} is a set of operators with different connection topologies, a \textit{stage} contains a repetition of the same block type and the \textit{backbone} determines how we connect blocks to form distinct networks. 

\subsubsection{The HiAML family}
(Fig.~\ref{fig:families}(a)) first constructs a network backbone with 4 stages 
and exactly 2 block repetitions per stage. 
Different from other families, the block structures are not randomly sampled. 
Instead, we first run GA-NAS~\cite{rezaei2021generative} 
on a search space similar to NAS-Bench-101, i.e., we allow up to 4 operators per block with any connection patterns in-between. 
After running NAS we find 14 blocks, and to construct the HiAML family we uniformly sample from these 14 blocks for each stage.
Reduction is performed in the stem Conv3x3 and the first block of stages 2 and 4.

We label 4629 networks for this family on CIFAR-10 by training them for 200 epochs with a batch size of 256. We optimize the network using Stochastic Gradient Descent (SGD) with a Nesterov momentum coefficient of 0.9 and a weight decay of $1e^{-5}$. An initial learning rate of $1e^{-2}$ is annealed down to $1e^{-5}$ using a cosine schedule. Finally, we perform label smoothing with a value of $\epsilon = 0.1$.

\subsubsection{The Inception family}
(Fig.~\ref{fig:families}(b)) mimics the Inception-v4~\cite{szegedy2017inception} classification networks. Our Inception networks all contain 3 stages. 
Inside each stage
, we have 2-4 stacked blocks. 
Like HiAML, 
the blocks inside a stage 
have the same internal structure.  
Blocks in different stages will likely 
can have different internal structures. 
When sampling a block, one first chooses the number of branching paths. 

We allow up to 4 branching paths and inside each path, there are 
up 
to 4 operators. 
An operator could be a Conv3x3-BN-ReLU combination. 
We adopt a wide range of candidate operators, including regular Convolution, Depth-wise Separable Convolution, MBConv, etc. 
Figure~\ref{fig:families} (b) shows an example of a 2-path block topology. Additional paths will follow a similar layout. 
Note that we also divide the candidate operators into two groups: Conv operators (Conv op) and supplementary operators (Supp op). 

The supplementary operators set includes all 1x1 convolution and pooling operations, whereas the Conv operators set only includes convolution with kernel size $>$1. 
A path must have at least one 1 Conv op, and at most one Supp op. 
Input channels will be further reduced based on the number of paths, e.g., if there are 2 paths then the channel size of a path will be half of the input channels, achieved via projection by the first operator of the path. 
In our Inception networks, we perform reductions on the input image in each stage by halving the height and width of the input and doubling the channels in the first block of each stage. %. This is achieved by the first block of each stage. 

We label 580 networks for this family on CIFAR-10. We train Inception networks for 500 epochs with a batch size of 256. We perform optimization using Stochastic Gradient Descent (SGD) with a Nesterov momentum coefficient of 0.9 as well as an initial learning rate of $1e^{-2}$ that we anneal down to $1e^{-4}$ using a cosine schedule. Like HiAML, we use label smoothing with $\epsilon = 0.1$.

\subsubsection{The Two-Path family}
\begin{figure*}[t!]
\begin{center}
\includegraphics[width=5.95in]{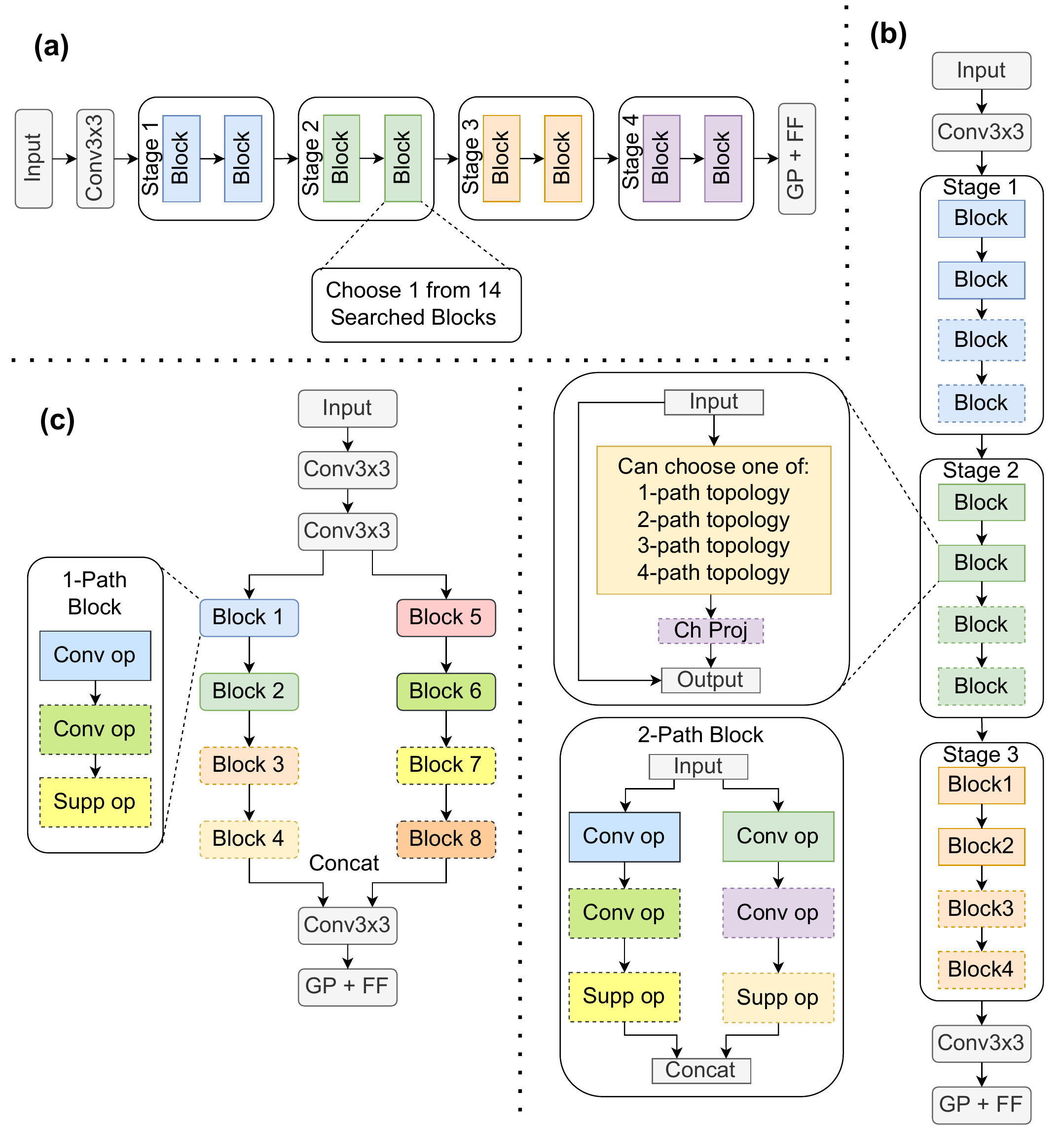}
\end{center}
\caption{
Illustrations of network and block design spaces for our proposed new families: (a) HiAML, (b) Inception and (c) Two-Path. 
For Inception and HiAML networks, each stage 
contains the repetitions of one type of block. 
Optional operators or blocks have dashed boundary lines. 
GP + FF stands for global average pooling and feedforward layers.
}
\vspace{-2mm}
\label{fig:families}
\end{figure*}

(Fig.~\ref{fig:families}(c)) constructs a 
simpler 
structure where an entire network contains only two branching paths. 
Inside a path, there are 2-4 blocks.  
Each block 
has a single-path topology with 1-3 operators. 
The definition of an operator is the same as the Inception family. In fact, one may view the Two-path family as the complement of the Inception family. An Inception network has a single-path backbone with branching paths inside blocks. But a Two-Path network has two branching paths of blocks, while each block is a single path of operators.
Reduction on the input image is performed in the second stem Conv3x3 
and also by two randomly chosen blocks of each path.
Since we do not allow block repetitions, Two-Path networks are usually lightweight and shallower 
than the other families. 
We label 6890 networks for this family on CIFAR-10 using the same hyperparameter setup as HiAML, e.g., 200 epochs.

\begin{table*}[t]
    \centering
    \scalebox{0.8}{
    \begin{tabular}{l|c|c|c|c|c|c|c} \toprule
    \textbf{Family} & \textbf{\textit{k}-GNN} & \textbf{CL} & \textbf{CL+T }& \textbf{CL+FCM+T} & \textbf{CL+Pairwise} & \textbf{CL+FCM+Pairwise} & \textbf{GENNAPE} \\ \midrule
    NB-201 & 0.4930 & \textbf{0.8520} & 0.7834 & 0.7647 & 0.8151 & \textit{0.8503} & 0.8146 \\
    w/ FT & 0.8606 $\pm$ 0.0245 & 0.7892$\pm$0.0715 & 0.8309$\pm$0.0410 & 0.8281$\pm$0.0181 & \textit{0.8897$\pm$0.0153} & 0.8896$\pm$0.0179 & \textbf{0.9103$\pm$0.0114} \\
    NB-301 & {0.0642} & -0.5250 & -0.1689 & \textit{0.1349} & -0.0702 & -0.2462 & \textbf{0.3214} \\
    w/ FT & \textit{0.8584$\pm$0.0290} & 0.6354$\pm$0.0929 & 0.7665$\pm$0.0318 & 0.7227$\pm$0.0275 & 0.8301$\pm$0.0335 & 0.8445$\pm$0.0093 & \textbf{0.8825$\pm$0.0134} \\\midrule
    PN & 0.0703 & 0.0138 & \textbf{0.8782} & 0.8068 & 0.4424 & 0.3489 & \textit{0.8213} \\
    w/ FT & 0.7559$\pm$0.0621 & 0.9209$\pm$0.0009 & 0.9281$\pm$0.0028 & 0.9322$\pm$0.0078 & 0.9429$\pm$0.0010 & \textit{0.9447$\pm$0.0064} & \textbf{0.9506$\pm$0.0039} \\
    OFA-MBv3 & 0.4345 & 0.5205 & 0.7122 & \textbf{0.8713} & 0.5350 & 0.5515 & \textit{0.8660} \\
    w/ FT & 0.6862$\pm$0.0253 & 0.9182$\pm$0.0070 & 0.9302$\pm$0.0021 & 0.9264$\pm$0.0050 & 0.9356$\pm$0.0062 & \textit{0.9369$\pm$0.0080} & \textbf{0.9449$\pm$0.0015} \\
    OFA-RN & \textbf{0.5721} & -0.6963 & -0.6035 & -0.7766 & -0.3541 & -0.5278 & \textit{0.5115} \\
    w/ FT & \textit{0.9102$\pm$0.0146} & 0.8588$\pm$0.0133 & 0.8187$\pm$0.0367 & 0.8350$\pm$0.0283 & 0.8785$\pm$0.0207 & 0.8737$\pm$0.0132 & \textbf{0.9114$\pm$0.0063} \\\midrule
    HiAML & -0.1211 & \textit{0.4373} & 0.4043 & \textbf{0.4511} & 0.3958 & 0.4167 & 0.4331 \\
    w/ FT & \textbf{0.4300$\pm$0.0507} & 0.2259$\pm$0.1753 & 0.2706$\pm$0.0805 & 0.2779$\pm$0.0545 & 0.3062$\pm$0.1187 & 0.3266$\pm$0.0701 & \textit{0.4169$\pm$0.0479} \\
    Inception & -0.2045 & 0.1843 & 0.3439 & \textbf{0.4429} & 0.3038 & 0.3487 & \textit{0.4249} \\
    w/ FT & 0.3340$\pm$0.0793 & 0.4771$\pm$0.0639 & \textit{0.5144$\pm$0.0448} & 0.4569$\pm$0.0547 & 0.3554$\pm$0.0759 & 0.3654$\pm$0.0825 & \textbf{0.5524$\pm$0.0166} \\
    Two-Path & 0.1970 & 0.0106 & \textit{0.3205} & {0.3130} & 0.1843 & 0.2866 & \textbf{0.3413} \\
    w/ FT & 0.3694$\pm$0.0406 & 0.3046$\pm$0.0977 & \textit{0.4412$\pm$0.0591} & 0.4331$\pm$0.0584 & 0.3251$\pm$0.1179 & 0.3432$\pm$0.0924 & \textbf{0.4875$\pm$0.0311} \\ 
    \bottomrule
    \end{tabular}
    }
    \caption{Spearman's rank correlation (SRCC) results across test families and predictors in the zero-shot transfer and fine-tuning (w/ FT) contexts. Higher is better. Best and second-best entries for each family are in bold and italics, respectively. Fine-tuning results averaged across 5 random seeds.}
    \label{tab:srcc_ablate}
\end{table*}

\begin{table*}[t]
    \centering
    \scalebox{0.8}{
    \begin{threeparttable}[hb]
    \begin{tabular}{l|c|c|c|c|c|c|c} \toprule
    \textbf{Family} & \textbf{\textit{k}-GNN} & \textbf{CL} & \textbf{CL+T }& \textbf{CL+FCM+T} & \textbf{CL+Pairwise} & \textbf{CL+FCM+Pairwise} & \textbf{GENNAPE} \\ \midrule
    NB-201 & 0.3382 & \textit{0.6554} & 0.5855 & 0.5719 & 0.6253 & \textbf{0.6572} & 0.6202 \\
    w/ FT & 0.6772$\pm$0.0100 & 0.6004$\pm$0.0739 & 0.6430$\pm$0.0445 & 0.6376$\pm$0.0205 & \textit{0.7137$\pm$0.0209} & 0.7121$\pm$0.0244 & \textbf{0.7415$\pm$0.0180} \\
    NB-301 & 0.0431 & -0.3670 & -0.1122 & \textit{0.0921} & -0.0484 & -0.1673 & \textbf{0.2183}\\
    w/ FT & \textbf{0.7125$\pm$0.0143} & 0.4644$\pm$0.0753 & 0.5767$\pm$0.0315 & 0.5340$\pm$0.0257 & 0.6474$\pm$0.0346 & 0.6614$\pm$0.0100 & \textit{0.7075$\pm$0.0171} \\\midrule
    PN & 0.0473 & 0.0095 & \textbf{0.6999} & 0.6174 & 0.3057 & 0.2375 & \textit{0.6299} \\
    w/ FT & 0.5927$\pm$0.0421 & 0.7593$\pm$0.0137 & 0.7711$\pm$0.0034 & 0.7796$\pm$0.0115 & 0.7978$\pm$0.0176 & \textit{0.8006$\pm$0.0122} & \textbf{0.8135$\pm$0.0074} \\
    OFA-MBv3 & 0.2995 & 0.3637 & 0.5221 & \textbf{0.6892} & 0.3737 & 0.3900 & \textit{0.6832} \\
    w/ FT & 0.5016$\pm$0.0128 & 0.7564$\pm$0.0114 & 0.7754$\pm$0.0036 & 0.7689$\pm$0.0082 & 0.7843$\pm$0.0112 & \textit{0.7868$\pm$0.0146} & \textbf{0.8024$\pm$0.0027} \\
    OFA-RN & \textbf{0.4095} & -0.5054 & -0.4320 & -0.5858 & -0.2565 & -0.3549 & \textit{0.3559} \\
    w/ FT & \textit{0.7437$\pm$0.0139} & 0.6760$\pm$0.0155 & 0.6302$\pm$0.0374 & 0.6466$\pm$0.0327 & 0.6997$\pm$0.0247 & 0.6897$\pm$0.0139 & \textbf{0.7457$\pm$0.0092} \\\midrule
    HiAML & -0.0813 & 0.3030 & 0.2809 & \textbf{0.3166} & 0.2726 & 0.2880 & \textit{0.3040} \\
    w/ FT & \textit{0.2875$\pm$0.0581} & 0.1629$\pm$0.1195 & 0.1940$\pm$0.0539 & 0.1988$\pm$0.0360 & 0.2104$\pm$0.0828 & 0.2240$\pm$0.0488 & \textbf{0.2913$\pm$0.0352} \\
    Inception & -0.1383 & 0.1244 & 0.2356 & \textbf{0.3059} & 0.2047 & 0.2400 & \textit{0.2935} \\
    w/ FT & 0.2082$\pm$0.0352 & 0.3335$\pm$0.0453 & \textit{0.3627$\pm$0.0337} & 0.3212$\pm$0.0404 & 0.2424$\pm$0.0544 & 0.2516$\pm$0.0572 & \textbf{0.3911$\pm$0.0137} \\
    Two-Path & 0.1397 & 0.0753 & \textit{0.2314} & 0.2257 & 0.1306 & 0.2047 & \textbf{0.2323} \\
    w/ FT & 0.2232$\pm$0.0432 & 0.2115$\pm$0.0670 & \textit{0.3094$\pm$0.0430} & 0.3033$\pm$0.0420 & 0.2222$\pm$0.0824 & 0.2342$\pm$0.6370 & \textbf{0.3389$\pm$0.0230} \\\bottomrule
    \end{tabular}
    \end{threeparttable}
    }
    \caption{Kendall's Tau (KT) results across test families and predictors in the zero-shot transfer and fine-tuning (w/ FT) contexts. Higher is better. Best and second-best entries for each family are in bold and italics, respectively. Fine-tuning results averaged across 5 random seeds.}
    \label{tab:kendall_ablate}
    \vspace{-1mm}
\end{table*}

\subsection{Model and Hyperparameter Details}

\subsubsection{Baseline GNNs}
\label{sec:gnnDetails}
Our GNNs 
(e.g., $k$-GNN, as well as GCN~\cite{welling2016semi} and GIN~\cite{xu2019GIN}) consist of an initial embedding layer that transforms CG node features into continuous vectors of size 32. Graph encoders consist of 6 layers. We derive the graph embedding by averaging all node embeddings. Regressors consist of 4 layers with a hidden size of 32. Baseline GNNs train predictors for 40 epochs with an initial learning rate of $1e^{-4}$ and a batch size of 32.

\subsubsection{Contrastive Learning Graph Encoder}
\label{sec:clDetails}
The CL encoder
consists of a GNN~\cite{morris2019weisfeiler} and Transformer~\cite{vaswani2017attention} whose output is concatenated to produce CG embeddings. The GNN consists of 4 layers, while the Transformer consists of a single layer with two attention heads trained with a dropout rate of 0.1. Both modules produce output vectors of length 64, so the overall embedding size is $m = 128$. We train the CL encoder using a batch size of 128, consider the $q = 21$ smallest eigenvalues and use a temperature $\tau = 0.05$. 
For regularization, we add a second projection head that predicts the FLOPs of a CG.

\subsubsection{Details of the Fuzzy C-Means Ensemble}
\label{sec:fcmDetails}

We use a simple grid search to determine the number of clusters $C \in [10, 20]$ and fuzzification coefficient $m \in [2.0, 4.0]$. As $m$ is continuous we consider increments of $0.5$ and select the model that achieves the highest SRCC the second NB-101 validation set. For regressors, these were $[C, m] = [16, 4.0]$ and for classifiers it was $[C, m] = [11, 3.5]$. Prior to clustering, we standardize FLOPs using the mean and standard deviation from the NB-101 training set and concatenate them with the CL graph embeddings. We also use Principle Component Analysis (PCA)~\cite{wold1987principal} to reduce embedding size down to 32. FCM runs for 10k iterations or until the stopping criteria $\epsilon = 1e^{-9}$ is satisifed. 

\subsubsection{Details of the CL and CL+FCM}
\label{sec:mlpDetails}

MLP regressors contain 4 hidden layers with a hidden size of 256 each, while pairwise classifiers contain 128. FCM memberships in the classifier sum to produce a latent vector of size 16 for each CG in a pair. We concatenate these and use an MLP layer to produce a prediction. We use ReLU as an activation function and train all predictors for 40 epochs with an initial learning rate of $1e^{-4}$ and a batch size of 32. We perform fine-tuning for 100 epochs using 50 samples with a batch size of 1. Finally, we obtain the SRCC of pairwise classifiers using mergesort with the model as a comparater to generate a list of rankings. 

\subsection{Calculation of NDCG}
\label{sec:ndcg}

We provide a brief explanation on the calculation of Normalized Discounted Cumulative Gain (NDCG) for the top-$k$ best architectures in a population. 
Given lists of predictions and ground-truth accuracies, sort into descending order according to the prediction. Each prediction shall be given an index $1, 2, 3,..., N$ while the labels will be assigned a relevance score $rel_1, rel_2, rel_3,..., rel_N$. Following AceNAS~\cite{zhang2021acenas}, we assign relevancy by re-scaling accuracy labels into the range $[0, 20]$. We first compute the Discounted Cumulative Gain (DCG) as

\begin{equation}
    \centering
    \label{eq:dcg}
    DCG = \sum_{i=1}^{k}\dfrac{2^{rel_i}-1}{\texttt{log}_2(i+1)},
\end{equation}
where $k \leq N$ allows one to control how many items we consider. We then compute NDCG as

\begin{equation}
    \centering
    \label{eq:ndcg}
    NDCG = \dfrac{DCG}{IDCG},
\end{equation}
where $IDCG$ is $DCG$ under the ideal ordering, e.g., descending sort according to the ground truth where $rel_i \geq rel_j$ if $i < j$. 

\subsection{Search Algorithm Details}
\label{sec:search_details}

We pair our fine-tuned CL+FCM+T predictors with a simple local search algorithm which runs for 6 iterations. In each iteration, we collect the current top-$k$ best CGs based on predicted performance and FLOPs.
For each top-$k$ CG, we perform mutation by randomly replacing a sequence of primitive operations with a sub-graph of new operations, ensuring that previous HWC channel constraints are met.
Finally, we rank the CGs based on the predicted performance and collect the new top-$k$ best CGs for the next iteration. 
When searching on NB-101 and NB-201, we select the optimal architectures on CIFAR-10. For OFA-ResNet, we use the best architecture found by \citet{mills2021profiling}.

\begin{table*}[t]
    \centering
    \scalebox{0.8}{
    \begin{threeparttable}[hb]
    \begin{tabular}{l|c|c|c|c|c|c|c} \toprule
    \textbf{Family} & \textbf{\textit{k}-GNN} & \textbf{CL} & \textbf{CL+T }& \textbf{CL+FCM+T} & \textbf{CL+Pairwise} & \textbf{CL+FCM+Pairwise} & \textbf{GENNAPE} \\ \midrule
    \multicolumn{8}{c}{\textbf{NDCG@50}} \\ \midrule
    NB-201 & 0.9495 & 0.9848 & \textit{0.9859} & 0.9847 & 0.9793 & \textbf{0.9872} & 0.9836 \\
    w/ FT & 0.9819$\pm$0.0056 & 0.9434$\pm$0.0214 & 0.9822$\pm$0.0063 & 0.9848$\pm$0.0040 & \textbf{0.9916$\pm$0.0022} & \textit{0.9907$\pm$0.0052} & 0.9885$\pm$0.0025 \\
    NB-301 & 0.7331 & 0.4175 & 0.6311 & \textit{0.7930} & 0.7235 & 0.5218 & \textbf{0.8184}\\
    w/ FT & \textit{0.9725$\pm$0.0043} & 0.8912$\pm$0.0480 & 0.9085$\pm$0.0430 & 0.9179$\pm$0.0357 & 0.9302$\pm$0.0690 & 0.9664$\pm$0.0102 & \textbf{0.9764$\pm$0.0045} \\\midrule
    PN & 0.6296 & 0.5264 & \textbf{0.9637} & \textit{0.9343} & 0.8935 & 0.6982 & 0.9099 \\
    w/ FT & 0.9225$\pm$0.0063 & 0.9619$\pm$0.0031 & \textit{0.9771$\pm$0.0061} & 0.9755$\pm$0.0097 & 0.9757$\pm$0.0082 & 0.9769$\pm$0.0046 & \textbf{0.9836$\pm$0.0037} \\
    OFA-MBv3 & 0.8607 & 0.8893 & 0.9268 & \textit{0.9397} & 0.8422 & 0.9182 & \textbf{0.9433} \\
    w/ FT & 0.8854$\pm$0.0203 & 0.9653$\pm$0.0028 & \textit{0.9691$\pm$0.0018} & 0.9660$\pm$0.0053 & 0.9644$\pm$0.0186 & \textit{0.9691$\pm$0.0110} & \textbf{0.9778$\pm$0.0018} \\
    OFA-RN & \textbf{0.8789} & 0.3144 & 0.2167 & 0.3494 & 0.3583 & 0.4344 & \textit{0.7007} \\
    w/ FT & \textbf{0.9746$\pm$0.0073} & 0.9491$\pm$0.0210 & 0.9245$\pm$0.0252 & 0.9077$\pm$0.0295 & 0.9520$\pm$0.0096 & 0.9222$\pm$0.0422 & \textit{0.9531$\pm$0.0169} \\\midrule
    HiAML & 0.5097 & 0.7534 & \textbf{0.7681} & \textit{0.7657} & 0.7567 & 0.7470 & 0.7173 \\
    w/ FT & 0.7505$\pm$0.0380 & 0.6847$\pm$0.0893 & 0.6903$\pm$0.0550 & 0.7172$\pm$0.0411 & 0.7562$\pm$0.0369 & \textit{0.7570$\pm$0.0284} & \textbf{0.7938$\pm$0.0144} \\
    Inception & 0.6196 & 0.7943 & \textbf{0.8607} & 0.8270 & 0.8240 & 0.8386 & \textit{0.8391} \\
    w/ FT & 0.7929$\pm$0.0265 & \textbf{0.8515$\pm$0.0287} & 0.8230$\pm$0.0206 & 0.8201$\pm$0.0133 & 0.7843$\pm$0.0194 & 0.8019$\pm$0.0340 & \textit{0.8452$\pm$0.0089} \\
    Two-Path & 0.6930 & 0.6722 & \textit{0.7886} & 0.7654 & 0.7272 & 0.7392 & \textbf{0.8115} \\
    w/ FT & 0.7860$\pm$0.0268 & 0.7839$\pm$0.0436 & \textit{0.8334$\pm$0.0304} & 0.8326$\pm$0.0281 & 0.7921$\pm$0.0514 & 0.7888$\pm$0.0435 & \textbf{0.8454$\pm$0.0185} \\\midrule
    \multicolumn{8}{c}{\textbf{NDCG@10}} \\ \midrule
    NB-201 & 0.9270 & 0.9845 & 0.9856 & \textbf{0.9868} & 0.9755 & \textit{0.9858} & 0.9793 \\
    w/ FT & 0.9751$\pm$0.0082 & 0.9362$\pm$0.0298 & 0.9818$\pm$0.0063 & 0.9834$\pm$0.0054 & \textbf{0.9910$\pm$0.0023} & \textit{0.9902$\pm$0.0063} & 0.9855$\pm$0.0030 \\
    NB-301 & 0.5341 & 0.3278 & 0.6411 & \textit{0.7737} & 0.7260 & 0.4046 & \textbf{0.7885}\\
    w/ FT & \textit{0.9723$\pm$0.0134} & 0.8767$\pm$0.0674 & 0.8979$\pm$0.0518 & 0.8961$\pm$0.0710 & 0.9150$\pm$0.0883 & 0.9666$\pm$0.0102 & \textbf{0.9765$\pm$0.0081} \\\midrule
    PN & 0.4426 & 0.4668 & \textbf{0.9564} & 0.9003 & \textit{0.9326} & 0.4849 & 0.8736 \\
    w/ FT & 0.9287$\pm$0.0271 & 0.9424$\pm$0.0045 & \textit{0.9759$\pm$0.0075} & 0.9716$\pm$0.0144 & 0.9663$\pm$0.0172 & 0.9752$\pm$0.0073 & \textbf{0.9800$\pm$0.0057} \\
    OFA-MBv3 & 0.8464 & 0.9092 & \textit{0.9252} & \textbf{0.9270} & 0.8067 & 0.9079 & 0.9234 \\
    w/ FT & 0.8859$\pm$0.0536 & 0.9558$\pm$0.0046 & 0.9636$\pm$0.0070 & 0.9620$\pm$0.0100 & 0.9596$\pm$0.0253 & \textit{0.9697$\pm$0.0132} & \textbf{0.9838$\pm$0.0030} \\
    OFA-RN & \textbf{0.9470} & 0.2515 & 0.2018 & 0.3036 & 0.3066 & 0.3859 & \textit{0.6606} \\
    w/ FT & \textbf{0.9717$\pm$0.0090} & 0.9416$\pm$0.0213 & 0.9174$\pm$0.0333 & 0.8932$\pm$0.0321 & \textit{0.9465$\pm$0.0140} & 0.9129$\pm$0.0422 & 0.9463$\pm$0.0236 \\\midrule
    HiAML & 0.5088 & \textit{0.7745} & \textbf{0.7925} & 0.7197 & 0.7498 & 0.7288 & 0.6892 \\
    w/ FT & 0.7356$\pm$0.0371 & 0.6743$\pm$0.0848 & 0.6721$\pm$0.0703 & 0.7150$\pm$0.0519 & 0.7470$\pm$0.0449 & \textit{0.7527$\pm$0.0444} & \textbf{0.7804$\pm$0.0211} \\
    Inception & 0.6064 & 0.8241 & \textit{0.8666} & 0.7540 & \textbf{0.8736} & 0.7834 & 0.8150 \\
    w/ FT & 0.7310$\pm$0.0423 & \textbf{0.8283$\pm$0.0312} & 0.7808$\pm$0.0369 & 0.7605$\pm$0.0276 & 0.7113$\pm$0.0445 & 0.7510$\pm$0.0550 & \textit{0.8073$\pm$0.0072} \\
    Two-Path & 0.6339 & 0.6496 & \textit{0.7993} & 0.7944 & 0.7223 & 0.7278 & \textbf{0.8275} \\
    w/ FT & 0.7860$\pm$0.0268 & 0.7839$\pm$0.0436 & \textit{0.8334$\pm$0.0304} & 0.8326$\pm$0.0281 & 0.7921$\pm$0.0514 & 0.7888$\pm$0.0434 & \textbf{0.8392$\pm$0.0220} \\\bottomrule
    \end{tabular}
    \end{threeparttable}
    }
    \caption{Normalized Discounted Cumulative Gain (NDCG) results across test families and predictors in the zero-shot transfer and fine-tuning (w/ FT) contexts. We consider NDCG@$50$ in addition to NDCG@$10$. Higher is better. Best and second-best entries for each family are in bold and italics, respectively. Fine-tuning results averaged across 5 random seeds.}
    \label{tab:ndcg_ablate}
    \vspace{-1mm}
\end{table*}

\subsection{Ablation Studies}
\label{sec:ablate}

We ablate the performance of several GENNAPE components, specifically the CL, CL+T and CL+FCM+T regressors as well as the CL+Pairwise and CL+FCM+Pairwise classifiers. We consider the zero-shot transfer and fine-tuning contexts, for all target families across three different metrics. 

\subsubsection{Spearman's Rank Correlation}

Table~\ref{tab:srcc_ablate} ablates SRCC results across all families. Overall, our findings demonstrate the generalizablity of GENNAPE as it still attains the best correlation performance on many families. Sometimes CL+FCM+T may outperform the full GENNAPE on a family (e.g., OFA-MBv3, HiAML, Inception) in the zero-shot transfer context, yet reinforcing the efficacy of our FCM and MLP-E scheme.

\subsubsection{Kendall's Tau}
We consider Kendall's Tau (KT), which counts the number of concordant and disconcordant pairs in a set and can handle ties. Table~\ref{tab:kendall_ablate} summarizes our findings. GENNAPE clearly outperforms other configurations on this metric in many scenarios. We note that in comparison to SRCC results in Table~\ref{tab:srcc_ablate}, the best performing predictors for any family and context do not change much. This is because KT is a correlation metric like SRCC that considers all samples equally, yet because it calculates over $N^2$ pairs rather than $N$ samples, KT is %performance numbers are 
generally closer to 0.

\subsubsection{Normalized Discounted Cumulative Gain}
NDCG calculations involve selecting the top-$k$ samples, e.g., NDCG@$k$. Before, we considered NDCG@$10$, but now broaden the scope of our ablation results to include NDCG@$50$. Table~\ref{tab:ndcg_ablate} presents our findings. For both values of $k$, GENNAPE most frequently achieves the highest performance, specifically for NDCG@$50$. However, as we decrease $k$ to $10$, several individual CL-based predictors attain the best performance, specifically the ensemble-driven CL+FCM+T. 

\subsection{Hardware and Software Details}
\label{sec:hardware}

We perform our experiments on rack servers that use Intel Xeon Gold 6140 CPUs with 756GB of RAM running Ubuntu 20.04.4 LTS. Each server holds a set of 8 NVIDIA Tesla V100 GPUs with 32GB each. 

To create Compute Graphs from models we use \texttt{TensorFlow==1.15.0}, reproducing \texttt{PyTorch} models using \texttt{Keras} components when necessary. To train predictor models we use \texttt{PyTorch==1.8.1} as well as \texttt{PyTorch-Geometric}. Additionally, we use \texttt{PyTorch-Fast-Transformers} to implement the transformer module in the CL graph encoder. 